\documentclass[journal]{IEEEtran}

\ifCLASSOPTIONcompsoc
  \usepackage[nocompress]{cite}
\else
  \usepackage{cite}
\fi
\usepackage{bbm}
\usepackage{amsmath}
\usepackage[noend]{algpseudocode}
\usepackage[ruled,vlined,linesnumbered]{algorithm2e}
\algrenewcommand{\Return}{\State \textbf{return}\ }
\algnewcommand{\funccall}[1]{\textit{#1}}
\usepackage{amsfonts}
\usepackage{amsmath}
\DeclareMathOperator*{\argmax}{argmax} 

  {\color{red}}%
  {}

\ifCLASSINFOpdf
\else
\fi
     \usepackage{graphicx}  
     \usepackage{float}
     \usepackage{cases}
\usepackage{graphicx}
\usepackage{caption}
     \usepackage{subfigure}     
     \usepackage{color,soul}
             \usepackage[table, dvipsnames]{xcolor}
             
             \usepackage{multicol}

\begin{document}
%
\title{Ensemble Consensus Representation Deep Reinforcement Learning for Hybrid FSO/RF Communication Systems}
%
%
%
%

\author{Shagufta Henna,~\IEEEmembership{Senior Member~IEEE,}
               Abid Ali Minhas, ~\IEEEmembership{Senior Member~IEEE,} Muhammad Saeed Khan, Muhammad Shahid Iqbal
\IEEEcompsocitemizethanks{\IEEEcompsocthanksitem Shagufta Henna is with the Computing Department, Atlantic Technological University, Donegal, Ireland.\protect\\
E-mail: shaguftehnna@gmail.com
}
\IEEEcompsocitemizethanks{\IEEEcompsocthanksitem Abid Ali Minhas is with the Department of Computer Engineering, Al Yamamah University,
Riyadh, Saudi Arabia.\protect\\
E-mail: a minhas@yu.edu.sa
}
\IEEEcompsocitemizethanks{\IEEEcompsocthanksitem Muhammad Saeed Khan is with the Department of Electrical
Engineering, Lahore Leads University, Lahore, Pakistan.\protect\\
E-mail: drmsaeed1980@gmail.com
}
\IEEEcompsocitemizethanks{\IEEEcompsocthanksitem Muhammad
Shahid Iqbal is with the Department of Electrical engineering National university of Technology,Islamabad, Pakistan.\protect\\
E-mail: shahid.iqbal@nutech.edu.pk
}

\IEEEcompsocitemizethanks{\IEEEcompsocthanksitem 
}

\thanks{Manuscript received April 19, 2005; revised August 26, 2015.}}

%
%

\markboth{}%
{Shagufta Henna \MakeLowercase{\textit{et al.}}: Bare Advanced Demo of IEEEtran.cls for IEEE Computer Society Journals}
%



\IEEEtitleabstractindextext{%
\begin{abstract}
Hybrid FSO/RF system requires an efficient FSO and RF link switching to improve the system capacity by realizing the complementary benefits of both the links. The dynamics of network conditions, such as fog, dust, and sand storms compound the link switching problem and control complexity. To address this problem, we initiate the study of deep reinforcement learning (DRL) for link switching of hybrid FSO/RF systems. Specifically,  we focus on actor-critic called Actor/Critic-FSO/RF  and Deep-Q network (DQN) called DQN-FSO/RF for FSO/RF link switching under atmospheric turbulences. To formulate the problem, we define the state, action, and reward function of a hybrid FSO/RF system. DQN-FSO/RF frequently updates the deployed policy that interacts with the environment in a hybrid FSO/RF system, resulting in high switching costs. To overcome this, we lift this problem to ensemble consensus representation learning-based DRL called DQNEnsemble-FSO/RF. The proposed  DQNEnsemble-FSO/RF DRL approach uses consensus learned features based on an ensemble of asynchronous threads to update the deployed policy. Experimental results corroborate that DQNEnsemble-FSO/RF's consensus-learned features demonstrate better performance than Actor/Critic-FSO/RF, DQN-FSO/RF, and MyOpic while keeping the switching cost low.  The results provide interesting insights into the prediction of received signal strength indicator (RSSI) for FSO/RF link switching.

\end{abstract}

\begin{IEEEkeywords}
Hybrid FSO/RF, DQN, reinforcement learning, FSO/RF link switching, actor-critic hybrid FSO/RF, representation learning for optical communication
\end{IEEEkeywords}}

\maketitle

\IEEEdisplaynontitleabstractindextext

%
\IEEEpeerreviewmaketitle

\ifCLASSOPTIONcompsoc
\IEEEraisesectionheading{\section{Introduction}\label{sec:introduction}}
\else
\section{Introduction}
\label{sec:introduction}
\fi

\IEEEPARstart{F}ree space optical communication (FSO) supports high data rate applications with minimum electromagnetic interference \cite{Androutsos19}. FSO requires a point-to-point link between transmitter and receiver and is sensitive to atmospheric turbulences, such as fog, dust, and sand storms resulting in degradation of FSO link capacity \cite{Ghassemlooy18, Katsilieris17}.  To increase the reliability of terrestrial broadband links, a radio frequency (RF) link is integrated with FSO links to form a hybrid FSO/RF system. This system exploits the complementary nature of individual FSO and RF links concerning their environmental sensitivity. This rationale is well explained as two links exhibit complementary attributes to atmospheric and weather conditions; the FSO link is not very sensitive to rain, while RF links do not deteriorate the signal quality from fog, sand, and dust storms and are heavily attenuated due to rain \cite{Cheng2020,Rakia2015,Lei2018}. A hybrid FSO/RF maintains reliable communication due to an efficient switching between FSO and RF links under varying weather conditions, thereby improving the performance of the system as a whole.

Designing an efficient hybrid FSO/RF system with high link availability in a dynamic environment characterized by fog, dust, or sand is challenging; essentially decisions regarding the switching of the link between FSO and RF must be made immediately \cite{Shakir2018,Douik2016}. Towards efficient FSO and RF link switching, few works have been reported, such as switching mechanism based on weak and strong turbulence for a 1-hop FSO/RF system  \cite{Touati2016,Sharma2019a}, fuzzy logic-based  FSO/RF soft switching mechanism \cite{Abid2021}, predictive link switching mechanism using long short term memory (LSTM) \cite{Meng2019}, AdaBoostClassifier, DecisionTreeClassifier, RandomForestClassifier, and GradientBoostingClassifier \cite{Toth2018}. Other methods include threshold-based link adaptation technique \cite{Bag2018} to avoid frequent link switching, FSO link outage-initiated link switching \cite{usman2014}, switching for multi-hop FSO/RF systems under various statistical models\cite{kong2015,Zedini2015, zhang2015}, and coding schemes for link switching \cite{Samimi2013,Ansari2013}. These techniques, although, switch FSO and RF links using threshold, coding, or predictive methods. However, these techniques are not efficient for link switching under a time-varying dynamic environment. The dynamics of network conditions certainly compound the problem and switching control complexity that is not addressed by the existing FSO/RF link switching techniques for hybrid FSO/RF systems.

Reinforcement learning (RL) enables an agent to learn a policy to maximize the expected sum of rewards. RL coupled with deep learning is known as deep reinforcement learning (DRL) learns difficult policies for complex problems, such as control tasks \cite{Babaeizadeh2017}. RL can efficiently learn state-action space under dynamic environments unknown for an agent. However, it is not suitable for scenarios with large state spaces with poor generalization capabilities. DQN efficiently solves these problems by using a deep neural network (DNN) to approximate a value function to select an optimal policy.

A RL agent can solve an optimization problem using an on-policy and off-policy. On-policy RL results in high sampling complexity where a policy acting on an environment and one being trained is the same, thereby requiring new samples for each training iteration. In contrast to on-policy, off-policy RL, such as DQN \cite{Mnih15}  exploits experience rather than relying only on new samples. Off-policy RL coupled with the neural networks for function approximation often leads to unstable behavior for continuous action space that is common in control/switching problems. An actor-critic DRL method addresses this instability issue with some actions using an intersection of off-policy, such as DQN, and policy-based methods, such as reinforcement learning \cite{Thomas017}.

Recent years have witnessed the application of artificial intelligence, such as DRL to various applications, such as efficient resource management and optimization \cite{Challita18,Fang18,Wei18,Ye19,Xiao18, hyang20},  fog/edge computing and caching \cite{Peng19,He17}, and dynamic channel assignment \cite{Wang18}. In \cite{yang2020}, authors introduced a heterogeneous hybrid visible light communication/RF system for industrial networks. The system targets to meet various quality of service (QoS) requirements, such as reliability, low latency communication, and higher data rates. The work also proposed a deep post-decision state-based experience replay method to support energy-efficient resource management. The learning algorithm learns an optimal policy with accelerated learning speed and efficiency.

To our best knowledge, there is no work to consider DRL for FSO and RF link switching or control problems in hybrid FSO/RF systems. This work proposes DRL-based frameworks for link switching in a hybrid FSO/RF system such as actor-critic and Deep-Q network (DQN). Due to frequently changing weather conditions, such as fog, dust, and sand, integrating FSO/RF system with efficient link switching is indeed challenging.  The key motivation for this DRL-based framework is to support dynamic and multi-factor decision making \cite{Luong19,Zhang20}.

DQN has gained significant attention due to tremendous improvement in performance with the help of deep neural networks to extract features. Recently, significant optimizations to DQN have been proposed, such as double DQN \cite{Hasselt10} to reduce overestimation of action values, experience replay buffer to prioritize learning \cite{Schaul15}, distributed DQN to model the distribution of action value, and dueling DQN architecture \cite{Wangz16} that maps action values from state and advantage values. 

To enable DQN efficiency for link switching and control decisions, it is desirable to limit a DQN agent to reduce the number of times the deployed/target policy changes during training. For large-scale hybrid FSO/RF systems, updating a policy requires reconsidering the complete environment. This requirement motivates the design of DQN agents with low switching costs. The proposed agent in this work aims to reduce the number of times the deployed/target policy interacts with the environment changes. Our work proposes consensus-based feature selection criteria to DQN to reduce switching costs in contrast to DQN. The key contributions of this work are summarized as follows:

\begin{itemize}

\item Conduct the first systematic study of modern deep reinforcement learning algorithms for hybrid FSO/RF systems. We implement two well-known DRL algorithms: 1) actor-critic based on actor and critic neural networks and 2) deep Q-network that consists of two deep neural networks to approximate a Q-function. We investigate these DRL methods to find the best policy for FSO and RF switching in contrast to MyOpic. This is illustrated below:

\begin{itemize}

\item  An actor-critic DRL-based method called Actor/Critic-FSO/RF solves the link switching optimization problem by considering its non-convex and combinatorial characteristics. The objective is to investigate the optimal long-term reward/utility of FSO and RF link switching for a hybrid FSO/RF system while maximizing link availability. Specifically, the states, action, and reward functions are defined for FSO/RF transceivers.

\item Adopt a deep Q-learning method and design called DQN-FSO/RF 
that can generate efficient FSO/RF link switching decisions in a highly dynamic environment subject to fog, rain, snow, and dust. With the help of a deep neural network,  DQN-FSO/RF  demonstrates a near-optimal strategy through approximation of action-value functions using the current state for a hybrid FSO/RF.

\end{itemize}
\item  Propose a novel consensus representation learning approach using deep Q-network called DQNEnsemble-FSO/RF to update the deployed policy to achieve low switching cost in a hybrid FSO/RF system in contrast to DQN-FSO/RF's frequent policy updates. Inspired by the representation learning, the switching criteria uses consensus features from an ensemble of asynchronous threads between the deployed Q-network and the underlying learning Q-network. The experiment results reveal that the proposed DQNEnsemble-FSO/RF  significantly reduces the policy switching cost in contrast to DQN-FSO/RF  and MyOpic when implemented on a hybrid FSO/RF system.

\item    Actor/Critic-FSO/RF  and DQN-FSO/RF strategies for FSO and RF link switching are evaluated under an environment subject to attenuations due to fog, dust, rain, and sand storms. Results exhibit the training performance of these DRL strategies under different parameters and show the influence of link switching on FSO/RF system performance in terms of reward.

\item  Experiment results show that the proposed DQNEnsemble-FSO/RF
DRL method demonstrates faster convergence and learning performance in contrast to Actor/Critic-FSO/RF and DQN-FSO/RF with a significantly reduced switching cost.

\item Experiments under an FSO/RF environment with various visibility ranges of 5km to 30km show that the DQNEnsemble-FSO/RF agent effectively predicts the received signal strength indicator (RSSI) required for accurate FSO/RF link switching. For example, results show that for the 2.5km visibility range, the minimum absolute error (MAE) is 0.34dBm, and AE values lower than 0.7 contribute to 90\% of all predictions.


\end{itemize}

The remainder of this article unfolds as follows. Section \ref{rw} discusses recent works relevant to link switching techniques in hybrid FSO/RF systems. We describe the system model and problem formulation in Section \ref{model}, respectively. Deep reinforcement learning methods, i.e., Actor/Critic-FSO/RF  and  DQN-FSO/RF for the FSO/RF link switching are presented in Section \ref{drl}. Section \ref{problem2} formulates the policy switching problem considering the hybrid FSO/RF system. The proposed DRL method to solve the policy switching problem called DQNEnsemble-FSO/RF
 is presented in Section \ref{secswitching}. Performance evaluation including MyOpic policy switching, evaluation setup, results, and analysis of all the DRL methods are given in Section \ref{pe}. Finally, Section \ref{conc} concludes the paper with possible future directions.

\section{Related Work} \label{rw}
Hybrid FSO/RF systems have been widely discussed in the literature to improve the reliability of communication. Authors in \cite{Chowdhury20} proposed a hybrid FSO/RF system to act as a  capacity backhaul to support high traffic for 5G and beyond. In this work, the FSO acts as a primary backhaul and RF as a secondary. RF system is activated through a one-bit feedback signal by the receiver, but the system does not consider the real-time channel conditions closer to the transmitter.

A link switching technique called adaptive combining used signal-to-noise ratio (SNR) threshold to keep the FSO link active is proposed in \cite{Rakia15}. It activates the RF link, if the perceived SNR of the FSO link drops below the switching threshold, enabling diversity reception for simultaneous data transmission over both the links combined through maximal ratio combining (MRC). The system switches back to the FSO link upon acceptable link quality. The combining technique based on MRC, however, is subject to performance and complexity tradeoff \cite{Rakia15c}. 

Works in \cite{Touati2016,Sharma2019a,Sharma2019b} proposed a FSO/RF link switching mechanism based on atmospheric turbulences. The switching mechanism keeps one link active depending upon atmospheric conditions. A method in \cite{Touati2016,Sharma2019a} evaluated an FSO/RF switching mechanism according to weak and strong turbulence for a 1-hop FSO/RF hybrid system. Abid et al. in \cite{Abid2021} proposed a fuzzy logic-based hybrid FSO/RF soft switching mechanism to improve the reliability and capacity of FSO under atmospheric conditions, such as sand/dust. The system consists of fuzzy-inference-based FSO and RF subsystem controlled using a fuzzy inference switching mechanism. The system improves the performance of the hybrid system for bit error rate (BER), SNR, and system capacity.  However, the fuzzy logic inference is dependent on human expertise and requires extensive validation and verification.

In \cite{Renat2020}, Renat et al. proposed a hard switching method based on received signal strength identifier (RSSI) predictions. The method aims to increase the high availability of the optical link using an RF link under atmospheric turbulences. Authors have considered the use of machine learning models, such as random forest, gradient-boosting regression, and decision trees for RSSI prediction to increase the availability of FSO/RF systems. Although, these models can accurately determine the predicted RSSI value. However, are these models are prone to over-fitting, thereby compromising their reliability. Further, decision trees are susceptible to small changes in data resulting in different outputs.

A work in \cite{Meng2019} proposed a predictive link switching mechanism for hybrid FSO/RF systems to conserve energy. The proposed method kept the FSO link continuously active and sampled the FSO signal periodically to collect a dataset. The method trained an LSTM model using the dataset. The work correlated the number of FSO signal power samples in the RF transmission with the prediction error using a predefined error threshold to improve energy efficiency.  This work, however, is preliminary and does not consider the dynamics in a hybrid FSO/RF communication system, a challenging task to the model environment to optimize network performance.

Another work in \cite{Toth2018} proposed a hard FSO/RF switching mechanism by predicting RSSI value. The authors analyzed the effects of the atmospheric channel on the quality of the optical signal. Although this analysis studied both the soft and hard switching between FSO and RF, the primary consideration was given to the hard switching. The work evaluated various machine learning classifiers, such as AdaBoostClassifier, DecisionTreeClassifier, RandomForestClassifier, and GradientBoostingClassifier for RSSI prediction to enable efficient hard FSO/RF switching.
Similarly to \cite{Meng2019}, the work is just an evaluation study of single/ensemble machine learning classifiers for the hard switching between FSO and RF links. The contribution is limited  and pays no attention to modeling dynamics of the atmospheric channel and its effect on soft switching.

According to the above investigation, it is evident that there are only preliminary studies including ensemble/single machine learning for link switching in hybrid FSO/RF systems. There are only a few works to harness the potentials of deep learning, such as LSTM in the field of hybrid FSO/RF systems. To our best knowledge, there is no work reported to investigate the performance of deep reinforcement learning for FSO/RF hybrid systems with a focus on link switching. Considering the practicability of deep reinforcement learning in radio resource allocation and management \cite{Hussain20}, in this work we enable consensus-based DRL for FSO/RF link switching under various weather turbulences.


\section{System Model} \label{model}
In this section, we presents the link switching problem in which hybrid FSO/RF system consisting of FSO and RF links switches link and learn the link states. Below, we describe the system model in detail. 

\subsection{Link Switching Pattern} 
 The hybrid system is equipped with RF and FSO with two possible states: $Tx_{ready}$ and $Tx_{switch}$. If a link is in $Tx_{ready}$ state, it can successfully transmit data. On the other hand, a $Tx_{switch}$ state indicates an imminent switching, otherwise, it can result in failed transmission. We assume that the states of these links are dynamically switching based on the atomospheric conditions, such as dust. FSO links interacts with the dust particles under Mie scattering. Mie scattering occurs when the diameter of dust particles is same as wavelengths of scattered light. Scattering coefficient is calculated using the distance, link visibility, and wavelenght of the transmitted bea. The relationship between the visibility range and dust is given in \ref{atten1}.

\begin{equation} \label{atten1} 
V=7080 \times C^{-0.8}
\end{equation}
 
The threshold value of attenuation level is selected based on the visibility and is calculated according to (\ref{atten1}) and (\ref{attenuation}) \cite{Abid2021}. In (\ref{atten1}), $V$ represents visibility in KM, and $C$ as the concentration of dust $(g/m^{3})$.
 The relationship between dust particles concentrations and scattering coefficient is given in (\ref{atten2}) \cite{Abid2021}.

 \begin{equation} \label{atten2} 
\tau_{s} = \exp \Bigg[  \bigg( \frac{-3.91}{7080 \times C^{-0.8}}  \bigg)  \bigg(\frac{\lambda}{0.55} \bigg)^{q} \times 	R \Bigg] 
\end{equation} 

In (\ref{atten2}),$\tau_{s}$ is scattering transmission, $\lambda$ denotes wavelength of the signal, $q$ is constant, and $R$ as the propagation range.

In \cite{Abid2021}, authors investigated attenuation of the signal due to dust on FSO links using (\ref{atten1}). According to this work, the presence of atmospheric particles affects the visibility measured that correlates directly with the attenuation. The atmospheric visibility $V$, and wavelength of the beam $\lambda$ are used to calculate the specific attenuation $\gamma$ as given in (\ref{attenuation}).

\begin{equation} \label{attenuation}
\gamma(\lambda)= 10 \log_{10}(e) \frac{3.912}{V} \bigg(\frac{\lambda}{0.55}\bigg)^{-q(V)}
\end{equation}
$q(V)$ in (\ref{attenuation}) is calculated as given in (\ref{visibility}).

\begin{equation} \label{visibility}
q(V)=\\
\begin{cases}
  1.6 & V > 50\\
   1.3 & 6 < V < 50 \\
       0.58V^{1/3} & V < 6 \\
  \end{cases}
\end{equation}

In Equation  (\ref{visibility}), $V$ is measured in km and $\lambda$ in $\mu$m.

We have selected the range of $100dB/km$ to $120dB/km$ attenuation thresholds based on work in \cite{wali2018}. According to analysis of results given in \cite{wali2018} show that over 70\% of the time attenuation of FSO links approaches to $120dB/km$ based on Kruse and Kim model, thereby requiring link switched to RF. On the other hand, attenuation less than $100dB/km$ is considered too weak for optical communications, therefore RF link is used to increment the reward to train the DRL agent under attenuation values less than $100dB/km$.
 At any time slot $t$, the FSO/RF links can be in $Tx_{ready}$ and $Tx_{switch}$ states based on acceptable values of SNR as discussed in \cite{Abid2021}. FSO link is considered $Tx_{ready}$ if atmospheric observations/attenuation level denoted as $\gamma$ is less than $100dB/Km$.  If the value of $\gamma \geq 120dB/Km$, the link is switched to RF link due to attenuation of FSO link. We assume that the states of links are dynamically switching between $Tx_{ready}$ and $Tx_{switch}$ according to the attenuation levels. We can model the switching patterns of the link states as Markov chain $M$.

At any instant of time $t$, the link state is represented as  $Tx_{ready}^{FSO,t}$,   $Tx_{ready}^{RF,t}$,   $Tx_{switch}^{FSO,t}$, and   $Tx_{switch}^{RF,t}$. We assume that the link states only change at the start of each time slot and remains same within the time slot. The probability that the link states change from the current state to a different state in $M$ is $p$; and $(1-p)$ if it stays the same.

 \subsection{Transceiver 's Observations}
 
 The model assumes that the link switching pattern is unknown. Transceiver learns link switching patterns from the observation of attenuation levels $\gamma$ based on weather dynamics using (\ref{atten1}) and (\ref{attenuation}) and are based on acceptable value of SNR as discussed in \cite{wali2018}. The system observations of the system in time slot $t$ is denoted as $O^{link}_{i,t}$, where $i$ denotes FSO and RF links selected for time slot $t$ $t = 1, 2, \ldots M$ given in (\ref{observlink}).

\begin{equation} \label{state}
link^{i,t}=\begin{cases}
   +1 & \text{if $\gamma<100dB/km$} \hspace{2mm} \& \hspace{2mm}\text{link $i$ is } FSO\\
   +1 & \text{if $\gamma\geq 120dB/km$} \hspace{2mm} \& \hspace{2mm}\text{link $i$ is } RF \\
   -1 & otherwise \\
  \end{cases}
\end{equation}



Now, the observation of link $i$ is

\begin{equation} \label{observlink}
O^{link}_{i,t}= \phi_{i,t} link^{i,t}=  \begin{cases}
    link^{i,t} & \text{if $i$ link selected in $t$}\\
  0 & \text{if $i$ link is not selected in $t$ }
  \end{cases}
\end{equation}

 The link $i$ represents FSO and RF links and $\phi_{i,t}$ is the indicator defined as:

\begin{equation} \label{indicator}
\phi_{i,t} =  \begin{cases}
    1 & \text{if $i$ link selected in time slot $t$}\\
  0 & \text{if $i$ link is not selected in $t$ }
  \end{cases}
\end{equation}

According to above equations, if link $i$ is not selected for transmission or its state is not known that indicates that the observation of the links as zero.


\subsection{Action Space}
The transceiver can select FSO or RF link in each time slot. We consider a discrete action space  $A= \{link_{FSO}, link_{RF}\}$, corresponds to selecting RF or FSO link.  Hence, in each time slot, the system selects an action from the action space $A$, i.e., access the corresponding RF or FSO link, and the condition of the selected link will be revealed.

\subsection{Link Switching Problem}

In this section, we formulate link switching problem using the $\gamma$ as discussed in Section \ref{model}. This work considers link switching pattern learning using actor-critic, DQN, and novel  ensemble consensus DRL approach to make link selection decisions. The agent make link access decision based on the $\gamma$, and updates the policy accordingly.

Let us define the reward as $r_{i,t}$ if the link $i$, i.e., FSO or RF in time slot $t$  as given in (\ref{rewardFSORF}) below.

 \begin{equation} \label{rewardFSORF}
 r_{i,t}=link^{i,t}=\\
\begin{cases}
   +1 & \text{if $\gamma<100$} \hspace{2mm} \& \hspace{2mm}\text{ $i$ is }FSO\\
   +1 & \text{if $\gamma\geq 120$} \hspace{2mm} \& \hspace{2mm}\text{ $i$ is } RF \\
        -1 & otherwise \\
  \end{cases}
\end{equation}
Since the hybrid system aims to select FSO or RF links based on  $\gamma$ in $dB/km$ to ensure successful transmissions, the system aims to extract a policy, i.e., to map observation $O^{link}$ to the action space $A$ to maximise the long-term expected reward $R$ of link access given in (\ref{utility}).

\begin{equation} \label{utility}
\pi^{*}= \operatorname*{arg\,max}_\pi R
\end{equation}

The reward $R$ for a finite number of time slots $T$ is expressed in (\ref{reward}) as:

\begin{equation} \label{reward}
\textstyle
R= \frac{1}{T}\sum_{t=1}^{T}\sum_{i=1}^{2}\phi_{i,t} r_{i,t}
\end{equation}
where $\phi_{i,t}$ is the indicator function defined in (\ref{indicator}). The problem can be formulated as:

\begin{equation} \label{problem}
 \hspace{4mm} \underset{\{\phi_{i,t}\}}   {Maximize}  \hspace{2mm}    R
\end{equation}
Subject to

\begin{equation*}
 \hspace{4mm} \sum_{i=1}^{2} \phi_{i,t}=k \hspace{2mm} \forall t
\end{equation*} 

where $k$ represents FSO and RF links that system can select in each time slot, and according to the
R, the $R \in [-k, k]$.



\section{Deep Reinforcement Learning-based Agent for Hybrid FSO/RF System} \label{drl}
This section presents the framework for the proposed hybrid FSO/RF system based on a deep reinforcement learning agent.  We consider various DRL approaches including DQN and actor-critic to switch between RF and FSO links. Further, we also consider the performance comparison of the actor-critic agent with the MyOpic policy under complex and dynamic weather conditions. Finally, we propose a consensus ensemble representation learning DQN approach for low switching costs under time-varying and dynamic weather conditions. To the best of our knowledge, this is the first study and implementation of DQN and actor-critic, and proposal of ensemble consensus representation learning DQN in the field of hybrid FSO/RF for link switching.

\subsection{Deep Reinforcement Learning Agent}

\begin{figure*}[htp]
\centering
\includegraphics[scale=0.58]{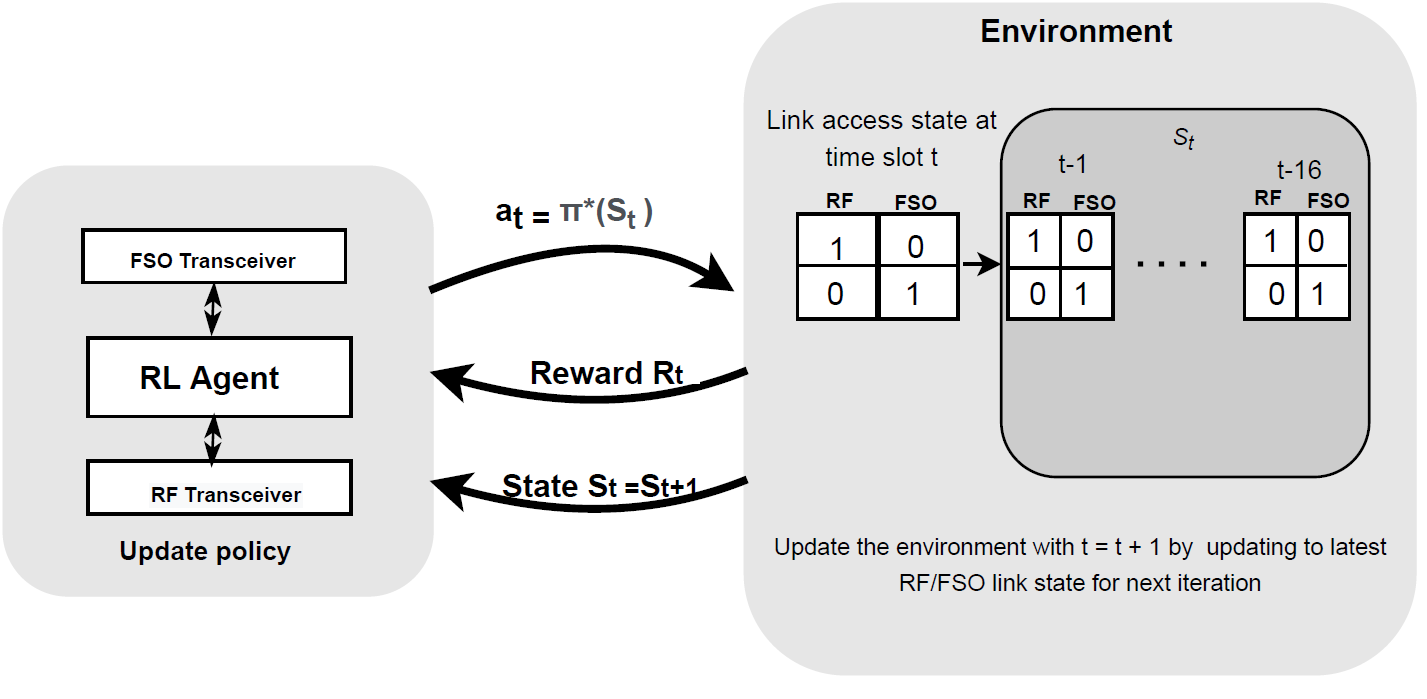}
\caption{Reinforcement learning agent for hybrid FSO/RF system.} \label{rlfso}
\end{figure*}

\textbf{FSO/RF link state:} As discussed earlier, the FSO/RF link state is time-varying and can be modeled by the Markov Decision Process (MDP). The agent uses its observation space as an input to the DRL framework and selects the FSO or RF link in each iteration. The agent updates the reward based on the state of the selected FSO or RF link. The FSO/RF system learns the best policy based on previous experiences in form of observation space for link $i$ as  $O^{link}_{i,t}$ as given in (\ref{observlink}) for time slot $t$ that is updated for time slot $t+1$ as $O^{link}_{(i,t+1)}=\{O^{link}_{t}, O^{link}_{t-1} \ldots O^{link}_{t-(MAX-1)}\}$, where MAX denotes the maximum number of iterations to observe the FSO/RF link state.

\textbf{Action space:} The agent evaluates different actions from the action space $A$ using the observation space $O^{link}_{i,t}$ with the highest reward. In the context of FSO/RF switching, an action means to access RF or FSO link for data transmission in time slot $t$.

\textbf{Reward:} Once the agent deliberate action from the $A$, it observes an instantaneous reward based $ r_{i,t}$ given in (\ref{rewardFSORF}) using $\gamma$. The basic objective of the reward is to solve the optimization problem as defined in (\ref{problem}).

We also define the reward received by the agent in time slot $t$ in addition to \ref{reward} for the actor-critic DRL agent as:

\begin{equation}\label{drlreward}
\textstyle
R_{t}= \sum_{i=1}^{2}\phi_{i,t}r_{i,t}
\end{equation}

where $\phi_{i,t}$ is the link selection indicator as given in (\ref{indicator}) and $r_{i,t}$ as in Equation \ref{rewardFSORF}.

\subsection{Actor/Critic-FSO/RF DRL Algorithm }	

\begin{figure*}[htp]
\centering
\includegraphics[scale=0.58]{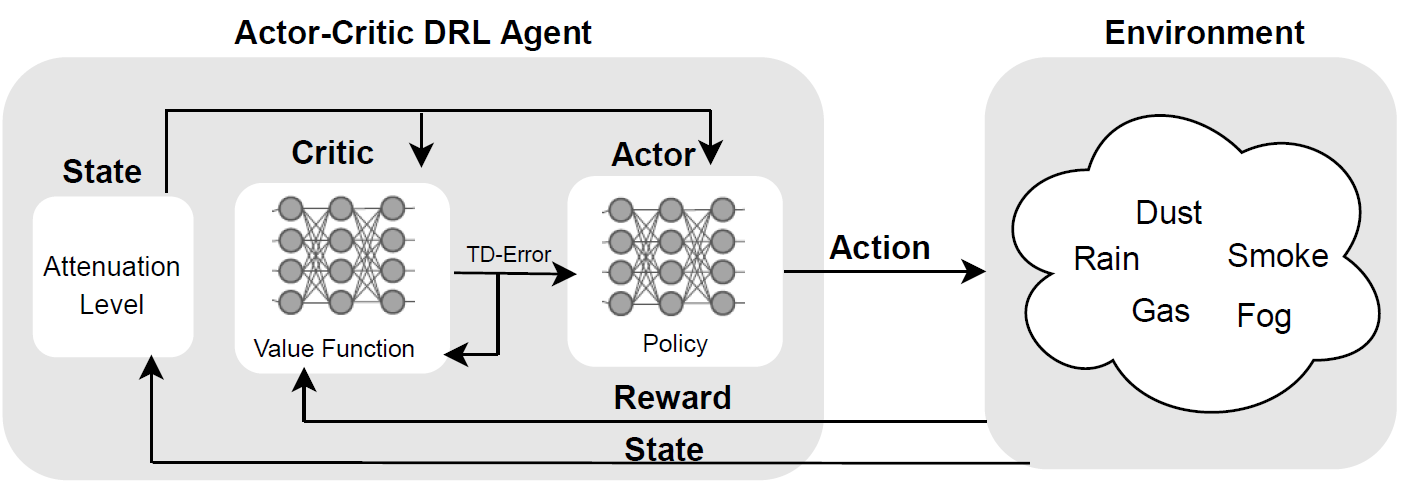}
\caption{Actor/Critic-FSO/RF agent for hybrid FSO/RF system.} \label{actorcritic}
\end{figure*}

It can be observed from Figure \ref{actorcritic} that the actor-critic framework is based on actor and critic neural networks. It can be observed from Algorithm \ref{alg:criticactor} that the actor neural network is initialized with $\theta$ and critic with parameter $\mu$. The actor neural network maps an observation $O^{link}$ at time slot $t$ to an actions $a$ from $A$  using the optimal policy $\pi^{*}$ as given in (\ref{actor1}). In the Equation, $A$ is discrete and the normalized probability of each action is calculated using the Softmax function at the output layer of the actor-network.

\begin{equation}\label{actor1}
\textstyle
\pi_{\theta}(O^{link}):O^{link} \rightarrow A
\end{equation}

In (\ref{actor1}), $\pi_{\theta}(O^{link})$ is the mapping policy which is a function of link observations $O^{link}$ parametrized by $\theta$ The actor neural network, therefore, can be represented as (\ref{actor2}).

\begin{equation}\label{actor2}
\textstyle
a= \pi_{\theta}(O^{link})
\end{equation}

The critic neural network is based on a value function, $V(O^{link})$. The critic neural network receives the feedback from the actor network that records the $V(O^{link})$ by  executing the action in the environment with varying weather conditions at time slot $t$. The actor feedback consists of $r_{t}$
and the observation $O^{link}_{t+1}$ for next time slot $t+1$. The critic uses this value to calculate the temporal difference error (TD) as given in (\ref{td}), where  $\gamma \in (0,1)$.

\begin{equation}\label{td}
\textstyle
\delta(t) \leftarrow R_{t}+\gamma V(O^{link}_{t+1})- V (O^{link}_{t}) 
\end{equation}

The critic network uses the optimal value function $Value^{*}$ to minimize the least squares temporal difference (LSTD) as given in (\ref{td1}).

  \begin{equation} \label{td1}
Value^{*}= \operatorname*{arg\,min}_{Value_{\mu}} (\delta(t))^2
\end{equation}

\begin{algorithm}
  \SetAlgoLined
 \KwData{ $\mu$, $\theta$}
  \KwResult{Reward}
  
  $critic_{FSO/RF} \leftarrow \mu$ 
  \Comment {Initialize the critic network to $\mu$}\\
  $actor_{FSO/RF} \leftarrow \theta$ 
  \Comment {Initialize the actor network to $\theta$} \\
  initialize the state of $FSO$ and $RF$  \\
$O^{link}_{t} \leftarrow 0$ 
 \Comment {Initialize the environment observations as zero}

         $i \leftarrow 0$
                  
           \While{$i \neq N$} {
           
           $a_{t} \leftarrow \pi (O_{i}|\theta)$
           \Comment{ The agent selects FSO/RF link as dictated by the policy under $\theta$} \\
         $R_{t} \leftarrow r_{(FSO/RF,t)}$ 
          \Comment{Access the selected link and receives reward based on the link state} 
                  $O^{link}_{t+1} \leftarrow O^{link}_{t}$ \\
               
 \For {next time slot $(O^{link}_{t+1}) $}{
$\delta_{t} \leftarrow R_{t}+\gamma V(O^{link}_{t+1})- V (O^{link}_{t}) $
\Comment{$critic_{FSO/RF}$ computes the TD error} \\
           $critic_{Loss}(O^{link}_{t}, a_{t}) \leftarrow \delta(t)^{2}$
           \Comment{minimize the $critic_{FSO/RF}$ loss} \\
            $\pi_{\theta}(O^{link}_{t} (\Delta \theta_{t} ) \leftarrow \alpha \nabla \theta_{t} log \pi \theta_{t} (O^{link}_{t}, a_{t}) $
            \Comment{Update actor policy to maximize value with $\alpha \in {0,1}$}
           
 $O^{link}_{t} \leftarrow O^{link}_{t+1}$  \\
           
          Update the $link^{FSO}_{t}$ and $link^{RF}_{t}$ state 
	  	             
	  	             }
	  	                             
                    }

\caption{Actor/Critic-FSO/RF Algorithm} \label{alg:criticactor}
\end{algorithm}
The actor uses the TD error given in (\ref{td}) to compute the policy gradient as given below in (\ref{grd1}).

 \begin{equation} \label{grd1}
 \nabla_{\theta}g(\theta) \leftarrow  T_{\pi_{\theta}} [ \nabla \theta_{t} log \pi \theta_{t} (O^{link}_{t}, a_{t}) \delta_{t}]
\end{equation}

In (\ref{grd1}), $pi \theta_{t} (O^{link}_{t}), a_{t}$ represents the score of action, i.e., selected FSO/RF link under the current optimal policy $\pi^{*}$. Given this the parameters of the actor, neural network can be updated using the $\Delta \theta_{t} \leftarrow \alpha \nabla \theta_{t} log \pi \theta_{t} (O^{link}_{t}, a_{t}) $ with learning rate of $\alpha \in {0,1}$. The gradient for the actor network is computed using the (\ref{grd2}).

 \begin{equation} \label{grd2}
 \theta_{(t+1)} \leftarrow \theta_{(t)} +  \alpha \nabla \theta_{t} log \pi \theta_{t} (O^{link}_{t}, a_{t}) \delta_{t}
\end{equation}

The critic network collects the $Max$ recent observations of FSO and RF links called as $O^{link}_{t}$  at the beginning of time slot $t$. The actor network chooses link $i$ based on the optimal policy $\pi^{*}$. Once the selected link is used for transmission, the observed reward is recorded. The critic network computes the TD error by using the reward, current observation state $O^{link}_{t}$, and the next observation state $O^{link}_{t+1}$. The computed error updates the gradients of both the critic and actor neural networks.

Algorithm \ref{alg:criticactor} presents the steps of the Actor/Critic-FSO/RF DRL agent. Line 1 and line 2 of the algorithm initialize two neural networks: actor and critic. Critic network $critic_{FSO/RF}$ is parametrized with $\mu$ and actor-network $actor_{FSO/RF}$ with $\theta$ 
In lines 1-2, Initial observations $O^{link}_{i, t}$, where $i$ is FSO or RF links for each time slot $t$ are initialized as $0$. Line 6 to line 8 selects the FSO/RF link for timeslot $t$ as dictated by the policy $\pi$ under $\theta$, receives the reward based on the link-state, and updates the observation for the next time slot $t$.In line 10 in Algorithm,  based on the observation $O^{link}_{t+1}$  for time slot $t+1$, the critic network computes the TD error. Line 11 uses an optimal function to minimize the error. In Line 12 in Algorithm \ref{alg:criticactor}, the actor-network chooses the RF or FSO link based on the optimal policy $\pi^{*}$.

\subsection{Deep-Q Network for Hybrid FSO/RF System (DQN-FSO/RF)}	
The following section illustrates link switching using DQN for the hybrid FSO/RF system.

\subsubsection{Q-learning}
Q-learning based hybrid FSO/RF system has the potential to learn switching patterns between FSO and RF links directly from $\gamma$ as discussed in the Section \ref{model}. This is an ideal solution for weather dynamics, such as dust storms. Q-learning for hybrid FSO/RF system aims to find an optimal policy for efficient FSO/RF switching to maximize the accumulated reward $R$ as defined in (\ref{reward}). Q-learning learns from actions outside the current policy called off-policy learning. The Q-value denoted as $Q(s,a)$ represents the cumulative reward of an agent being in state $s$ and performing the committed action $a$ and is given in (\ref{qvalue}).

 \begin{equation} \label{qvalue}
Q(s,a)= r(s,a)+  \beta max_{a} Q(s^{'},a)
\end{equation}

In (\ref{qvalue}), $Q(s,a)$ denotes a Q-function, $r(s,a)$ represents the instantaneous reward of a transmission on the selected FSO or RF link and the $max_{a} Q(s^{'},a)$ is the maximum Q-value for the next observation/state with a discount factor $\beta$. An agent uses Q-Table to store and look up Q-values representing future actions according to states. The expected Q-values are updated using the (\ref{updateQvalue}).

\begin{multline} \label{updateQvalue}
Q_{update}(s,a)= Q_{past}(s,a)+  \alpha ( r(s,a) \\ 
+\beta max_{a} Q(s^{'},a)-Q_{past}(s,a))
\end{multline}
Equation given (\ref{updateQvalue}) computes the expected future Q-value as a difference between the old Q-value and the discounted future value with one-step look ahead.

\subsubsection{DQN-FSO/RF}
Q-learning can find an optimal policy if it can estimate Q-function for each action and state pair using (\ref{qvalue}). This turns computationally expensive for large state and action space. DQN \cite{Luong19} uses a deep neural network called as Q-Network to approximate the Q-function, i.e., $Q(s,a)= r(s,a)+  \beta max_{a} Q(s^{'},a)$. The agent selects an action $a_{t}$ using $Q(s_{t},a)$ and makes transition to next state $s_{t+1}$ with reward $r_{t+1}$. It is represented as a transition/experience tuple $ Exp_{t}=(s_{t},a_{t},r_{t+1}, s_{t+1})$. The $Exp_{t}$ is saved to a replay memory buffer $D$.

\begin{figure*}[htp]
\centering
\includegraphics[scale=0.58]{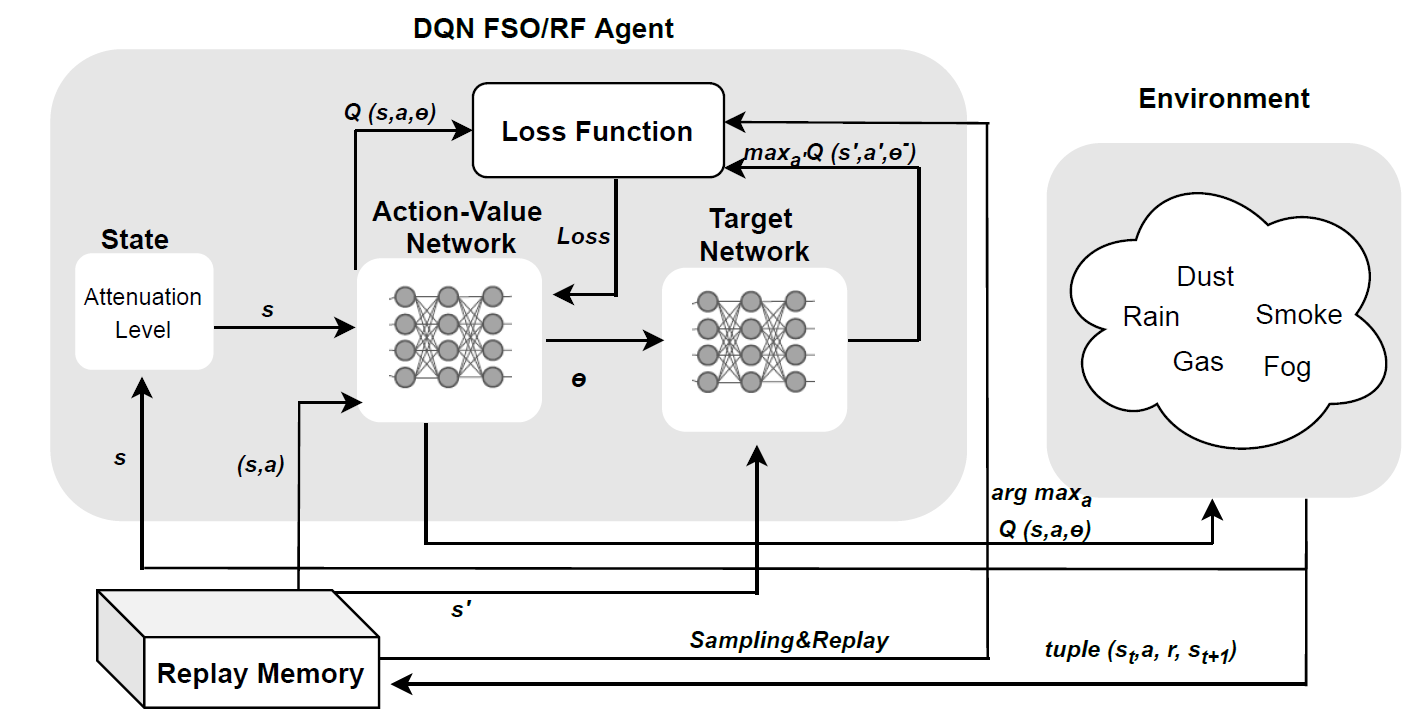}
\caption{DQN-FSO/RF agent for hybrid FSO/RF system.} \label{dqn}
\end{figure*}

DQN is based on states, actions, rewards, and the Q-Network $Q(s,a;\theta)$. At each time slot $t$ for link $i$, the Q-Network uses $O^{link}_{i, t}$ using (\ref{observlink}) to  select action $link_{FSO}$ or $link_{RF}$ from the action space $A$ as illustrated in Figure \ref{dqn}. The objective of the Q-Network is to maximize the reward as given in (\ref{reward}). 

A DQN agent consists of two deep neural networks to approximate Q-function which is
illustrated in Figure \ref{dqn}. One acts as an action-value function approximator $Q(s,a,\theta_{i})$ and the second as the target action-value approximator $\hat{Q}(s,a,\theta_{i}^{-})$, where $\theta_{i}$ and $\theta_{i}^{-}$ represents weights of the neural networks for iteration $i$. The weights of the first neural network, $\theta_{i}$ are updated using the mini-batch of random samples from the replay memory buffer $D$ at each learning iteration $i$.

The weights $\theta_{i}^{-}$ of the second network are updated using the stochastic gradient descent (SGD) and backward propagation algorithm that minimizes the mean-square error(MSE) as loss function. Referring to (\ref{updateQvalue}), the loss is calculated as (\ref{loss}), where $\theta$ represents the weights of the Q-Network.

\begin{equation} \label{loss}
Loss(\theta)= (r_{t}+ \beta max_{a} \hat{Q}(s^{i}_{t+1},a^{'}; \theta^{-}_{i})-Q(s_{t}^{i},a_{t}^{i};\theta_{i}))^{2}
\end{equation}
In (\ref{loss}), $\theta^{-}$ represents the parameters/weights of the target neural network that is replaced by the parameters $\theta$ of training Q-Network every $K$ time steps as can be seen in Figure \ref{dqn}. Deep Q-Network uses mini-batch data from the replay buffer $D$ to train the Q-Network. Instead of using an $\epsilon$-greedy approach, the experience replay component exploits stochastic prioritization to generate the probability of actions that helps the neural network to converge. The steps of DQN-FSO/RF are summarized in Algorithm \ref{alg:DQN}.

Algorithm \ref{alg:DQN} shows the steps of the proposed DQN-FSO/RF agent.  Line 1-5 are basic initializations for the DQN-FSO/RF agent. These initializations include replay buffer size, mini-batch size, Xavier weights, action-value, and the target network. Line 6-11 shows that for $T$ time slots, each $Agent_{FSO/RF}$ observes the state of FSO or RF link. The agent selects the best action, $link_{FSO}$ or $link_{RF}$ to maximize the Q-value of the action-value network. Line 13-14 records the reward by executing the selected action and updates the state for time slot $t+1$. Lines 15 to 20 record the tuple that is stored in the experience replay buffer. For given time slot $t$, the tuple includes current $s$, selected action, i.e., $link_{FSO}$ or $link_{RF}$, observed  reward $r$, and the state for the next time slot $t+1$. Line 18 samples a mini-batch $Batch_{mini}$ from the replay buffer $D$.  This $Batch_{mini}$ later in line 19 is used to compute the loss function using the SGD for action-value network weights $\theta$. Line 20 updates the target network weight $\theta^{-} $ to the  action-value network weights $\theta$ every $F$ steps.

\begin{algorithm}
  \SetAlgoLined
 \KwData{ $\alpha$, $\theta$, $\theta^{-}$, $T$, $\beta$, $X$, $U$, $F$}
  \KwResult{Reward}
   
  $D \leftarrow X$ 
  \Comment {Initialize replay buffer with capacity $X$}\\
  $Batch_{mini} \leftarrow B$ 
  \Comment {Initialize mini-batch with size $B$} \\
  $\theta^{-} \leftarrow U ([-(1/\sqrt(n)), 1/\sqrt(n)])$
  \Comment {Xavier weight initialization with random number using a uniform probability distribution $U$, $n$ number of inputs} \\
    $Q(s,a) \leftarrow \theta$
   \Comment {Initialize  action-value network with weights $\theta$} \\
   $\hat{Q}(s,a) \leftarrow \theta^{-}$
   \Comment {Initialize target network $\hat{Q}$ with weight random weights $\theta^{-}$} \\

 \For {( each time step $t \in T $)}{
Reset $link^{RF}_{t}$ and $link^{FSO}_{t}$
\Comment{Reset state of the RF and FSO transceiver} \\
   \For{each $j$ $Agent_{FSO/RF}$}{
   $s_{t}^{j} \leftarrow link^{FSO}_{t}/link^{RF}_{t}$ using (\ref{state}) 
   \Comment{Observe state for each FSO or RF link} \\
   $a_{t}^{j} \leftarrow Max (a \in \{a_{FSO},a_{RF}\},Q(s,a, \theta)) $ 
      \Comment{Choose FSO or RF link that maximizes the Q-value of the action-value network}
   
   }
$j \leftarrow a_{t}^{j}$ 
\Comment{FSO/RF hybrid agent take action}\\
 $r_{(FSO/RF,t)} \leftarrow r_{t}^{j}$
 \Comment{$Agent(FSO/RF)$ $j$ observes reward }\\
 $s_{t+1}^{j} \leftarrow s_{t}^{j}$
  \Comment{Update state $ s_{t+1}^{j}$ to  $s_{t}^{j}$  }\\
  
\For{each $j$ $Agent_{FSO/RF}$}{
   $s_{t+1}^{j} \leftarrow link^{FSO}_{t}/link^{RF}_{t}$ using (\ref{state}) \\
   $D_{j} \leftarrow (s_{t}^{j},a_{t}^{j},r_{t}^{j}, s_{t+1}^{j})$
   \Comment{Store tuple $(s_{t}^{j},a_{t}^{j},r_{t}^{j}, s_{t+1}^{j})$ into replay memory}
   
   $Batch_{mini} \leftarrow sampling (D_{j})$
      \Comment{Uniformly sample  $Batch_{mini}$ from replay memory $D_{j}$}
   
$Loss= (r_{t}^{j}+ \beta max_{a} \hat{Q}(s^{j}_{t+1},a^{'}; \theta^{-}_{j})-Q(s_{t}^{j},a_{t}^{j};\theta_{j}))^{2} $ 
       \Comment{Calculate loss using SGD with respect to $\theta$} 
  
   $\theta^{-}_{j}\leftarrow \theta_{j}$
    \Comment{Update target network weights $\theta^{-}_{j}$ to $\theta$ every $F$ steps  } 
   
   }
  	          
	  	             }

\caption{Deep Q-Network Algorithm (DQN-FSO/RF)} \label{alg:DQN}
\end{algorithm}

\section{ DQN-FSO/RF Policy Switching Cost Problem Formulation} \label{problem2}

This section focuses on the policy switching cost called as $link_{switch}^{RF/FSO}$. This cost is used to optimize the DQN-FSO/RF agent. The cost represents the frequency of changes in deployed policy $\pi$ in action-value network, i.e., $Q(s,a)$ in Algorithm \ref{alg:DQN} in $T$ episodes as given below in (\ref{cost}):

\begin{equation} \label{cost}
link_{switch}^{RF/FSO}= \sum_{t=1}^{T-1} \prod \{ \pi_{t} \neq \pi_{t+1} \}
\end{equation}

The objective of the DQN-FSO/RF agent is to learn an optimal policy $\pi^{*}$ with small $link_{switch}^{RF/FSO}$ cost.

\section{Ensemble Consensus Representation Deep Reinforcement Learning  DQNEnsemble-FSO/RF for FSO/RF link switching} \label{secswitching}

Inspired by the representation learning \cite{Subramanian2018}, We adopt the concept that DQN learns to extract informative features of the states of environments using the consensus of an ensemble of threads. Our proposed criterion tries to switch the deployed policy according to the consensus of features.
In the proposed approach to ensemble consensus to DQN-FSO/RF, $M$ asynchronous threads sample batch of data from the replay buffer and then extract features of all states to train both the action-value network and target Q-network.

For a state $s$, the extracted features by $M$ threads are denoted as $f_{m}(target,s)$, $f_{m}(action-value,s)$, respectively. The similarity score between $f_{m}(target,s)$, $f_{m}(action-value,s)$ for each thread $m$ on state $s$ is defined in (\ref{consens}) as:

\begin{equation} \label{consens}    
 sim(s) \leftarrow \frac{\langle f_{m}(target,s), f_{m}(action-value,s) \rangle}{\vert\vert f_{m}(target,s) \times f_{m}(action-value,s)\vert\vert  }
\end{equation}

The average similarity score for a batch of states  denoted as $B$ for each thread $m$ is given in  (\ref{avgscore}) as:

\begin{equation} \label{avgscore} 
avgSim(B_{s}) \leftarrow  \frac{\sum_{s \in B_{s}} sim(s)}{\vert\ vert SS \vert \vert}
\end{equation}

The Equation \ref{conscore} computes the consensus score of features using (\ref{avgscore}).

\begin{equation} \label{conscore} 
consensusSim(B_{s}) \leftarrow  \cap_{m=1}^{Asyn_{threads}} avgSim(B_{s}) 
\end{equation}

The representation learned ensemble consensus score updates the  target/deployed policy whenever the  $consensusSim(B_{s}) \leq \alpha$, where $\alpha \in [0,1]$ is learning rate. The DQNEnsemble-FSO/RF algorithm is illustrated in Algorithm \ref{alg:DQNensemble}.

\begin{algorithm}
  \SetAlgoLined
 \KwData{ $B_{size}$, $\alpha$, $M$, $T$,  $\theta$, $\theta^{-}$, $X$}
  \KwResult{Reward}
  $\alpha \leftarrow \left[0,1\right]$
   
  $D \leftarrow X$ 
  \Comment {Initialize replay buffer with capacity $X$}\\
  $B \leftarrow B_{size}$ 
  \Comment {Initialize mini-batch with size $B_{size}$} \\
  $Asyn_{threads} \leftarrow M$
  
   \For {( each time step $t \in T $)}{
    \For{each $j$ $Agent_{FSO/RF}$}{
   
   \For {( $m=1$ to $Asyn_{threads} $)}{
        $B_{m} \in D$      
       \Comment{Sample a batch of size $B_{size}$ from replay buffer} \\
        \For {( For each state $s$ in $S$)}{
       
       $f_{m}(target,s)  \leftarrow extractFeatures(B_{m},Q(s,a))$
       \Comment {Extract features from  $B_{m}$ for target-network}
       $ f_{m}(action-value,s) \leftarrow extractFeatures(B_{m},\hat{Q}(s,a))$
       \Comment {Extract features from  $B_{m}$ for action-value network}
       
       $ sim(s) \leftarrow \frac{\langle f_{m}(target,s), f_{m}(action-value,s) \rangle}{\vert\vert f_{m}(target,s) \times f_{m}(action-value,s)\vert\vert  }$ 
       
         \Comment{Feature similarity between action-value and target network on $s$ }
       $avgSim(B_{s}) \leftarrow  \frac{\sum_{s \in B_{s}} sim(s)}{\vert\ vert SS \vert \vert}$
  \Comment{Calculate average similarity score on batch of states}
      }

      }
        $consensusSim(B_{s}) \leftarrow  \cap_{m=1}^{Asyn_{threads}} avgSim(B_{s})$ 
  \Comment{Calculate ensemble consensus features} \\
       \If{$consensusSim(B_{s}) \leq \alpha$}{
          $\theta^{-}_{j}\leftarrow \theta_{j}$
          
    \Comment{Update target network weights $\theta^{-}_{j}$ to $\theta$ whenever $consensusSim(B_{s}) \leq \alpha$ }
    
    } 
    }
   
      }

\caption{ Ensemble Consensus Representation Deep Reinforcement Learning DQNEnsemble-FSO/RF for Policy Switching} \label{alg:DQNensemble}
\end{algorithm}

Lines 1-4 in Algorithm \ref{alg:DQNensemble} perform initializations of learning rate $\alpha$, replay buffer size, mini-batch size, and the number of asynchronous threads. Line 6-14 presents the steps to compute the average similarity score by $M$ asynchronous threads. The average similarity score in lines 11-12 is calculated using the features extracted from both the action-value and target networks for each state $s$ using the mini-batch sampled from the replay buffer $D$ as given in lines 9-11.
Line 15 calculates the ensemble consensus features $consensusSim()$ using the average similarity score $ avgSim()$. In contrast to $F$ steps for the DQN-FSO/RF, line 16-18 updates the target network weight $\theta^{-} $ to the  action-value network weights $\theta$, whenever $consensusSim() \leq \alpha$.

\section{Time Complexity of DQNEnsemble-FSO/RF } \label{complexity}
Let us assume that the DQN consists of $K$ layers. The size of input layer is $H$ and other layers as $m$. The time required to find the switching threshold for $L$ links is $\mathcal {O}(H \log L)$.
The time complexity of each $Agent_{FSO/RF}$ to switch among $L$ links for $T$ number of iterations is given as:

\begin{equation} \label{complexity}
\mathcal {O}(T(Hm+(K-1)m^{2}+H \log L)
\end{equation}

Let $M$ denotes the number of asynchronous threads in the DQNEnsemble-FSO/RF, and the time required for the feature extraction and feature similarity between the action-value and target network is $F$. The time complexity for the DQNEnsemble-FSO/RF for $S$  number of states is $\mathcal {O}(S\log F)$. The training time complexity for each $Agent_{FSO/RF}$ using DQNEnsemble-FSO/RF using $M$ threads is:

\begin{equation} \label{complexity}
\mathcal {O}(T(Hm+(K-1)m^{2}+H \log L+M(S \log F))
\end{equation}
\section{Performance Evaluation} \label{pe}
This section presents and discusses (i) the effectiveness of $Agent_{FSO/RF}$ using deep reinforcement learning algorithms including DQN and actor-critic and (ii) the comparison of DQNEnsemble-FSO/RF agent with the Actor/Critic-FSO/RF,  DQN-FSO/RF, and MyOpic as discussed below.

\subsection{MyOpic Policy for FSO/RF Link Switching}
The MyOpic policy only accumulates immediate reward obtained from transceiver switching without considering the future. MyOpic agent always selects a transceiver to maximize the immediate expected reward.

Given the state space of Markov chain be $S=\{Tx_{ready}^{i,t},Tx_{switch}^{i,t}\}$ where $i \in \{link^{RF}, link^{FSO}\}$ is a vector  $[Tx_{ready}^{FSO},Tx_{switch}^{RF}, Tx_{ready}^{RF},Tx_{switch}^{FSO}]$, where $Tx_{ready}^{FSO}/Tx_{ready}^{RF}$ and $Tx_{switch}^{RF}/Tx_{switch}^{FSO}$ denotes binary vector representing the state of FSO or RF links: $\gamma<100dB/Km$ and $\gamma\geq 120dB/Km$ ($1$), and otherwise ($0$) illustrated in (\ref{myopic}).

\begin{equation} \label{myopic}
\hat{a}_{FSO/RF}=\argmax_{k \in FSO,RF} \sum_{i=1}^{4} \omega s_{i} \mathbbm{1} (s_{ik}(t)=1)
\end{equation} 
In (\ref{myopic}), $\mathbbm{1}$ is an indicator function and $\omega s_{i}$ is the conditional probability of the hybrid FSO/RF system in state $s_{i}$ given past decisions/observations. The MyOpic policy follows a simple round-robin FSO and RF switching procedure and is not considered optimal \cite{Ahmad2009}.

\subsection{Evaluation Setup }
We have implemented the proposed DQNEnsemble-FSO/RF,  DQN-FSO/RF, and Actor/Critic-FSO/RF using Tensorflow with the DRL environment in OpenAI Gym framework \footnote{http://gym.openai.com/}.  The code for these DRL agents is available on the GitHub repository \footnote{https://github.com/shenna2017/FSORFagent}. For each iteration,  DRL agents are trained using $600$ to $2000$ episodes.  The DQN-FSO/RF and DQNEnsemble-FSO/RF agents are created using Keras functions. The neural network has 5 layers; $3$ hidden, one input, and one output layer. All three hidden layers consist of $300$, $200$, and $100$ neurons, respectively.  These layers are implemented using ReLu activation functions with the linear output layer.
For both the DQNEnsemble-FSO/RF and DQN-FSO/RF agents, actions are selected using the Boltzman policy \cite{Song2018}. Our evaluations consider the use of Adam optimizer to minimize the loss function given in (\ref{loss}). Other parameters including mini-batch size, learning rate, discount factor $\sigma$, and experience replay buffer size are given in Table \ref{tab1}.

\begin{table*}[ht]
		\renewcommand{\arraystretch}{1}%
		\centering
		\caption{DQN hyperparameters}
		\label{tab1}
		\begin{tabular}{ |p{3cm}|p{1.5cm}| } 
			\hline
			\rowcolor{lightgray}
			Hyper-parameters
			& Values
			\\\hline
			Batch size
			& 32
			\\\hline
				Activation function
			& ReLU
			\\\hline
				Learning rate
			& $10^{-4}$
			\\\hline
			Experience replay size
			 & 1,000,000
			\\\hline	
			$\gamma$
			 & 0.9
			\\\hline				
					
		\end{tabular}
		
	\end{table*}



\subsection{Results and Analysis}
\begin{figure}[ht] 
    \includegraphics[width=.55\textwidth]{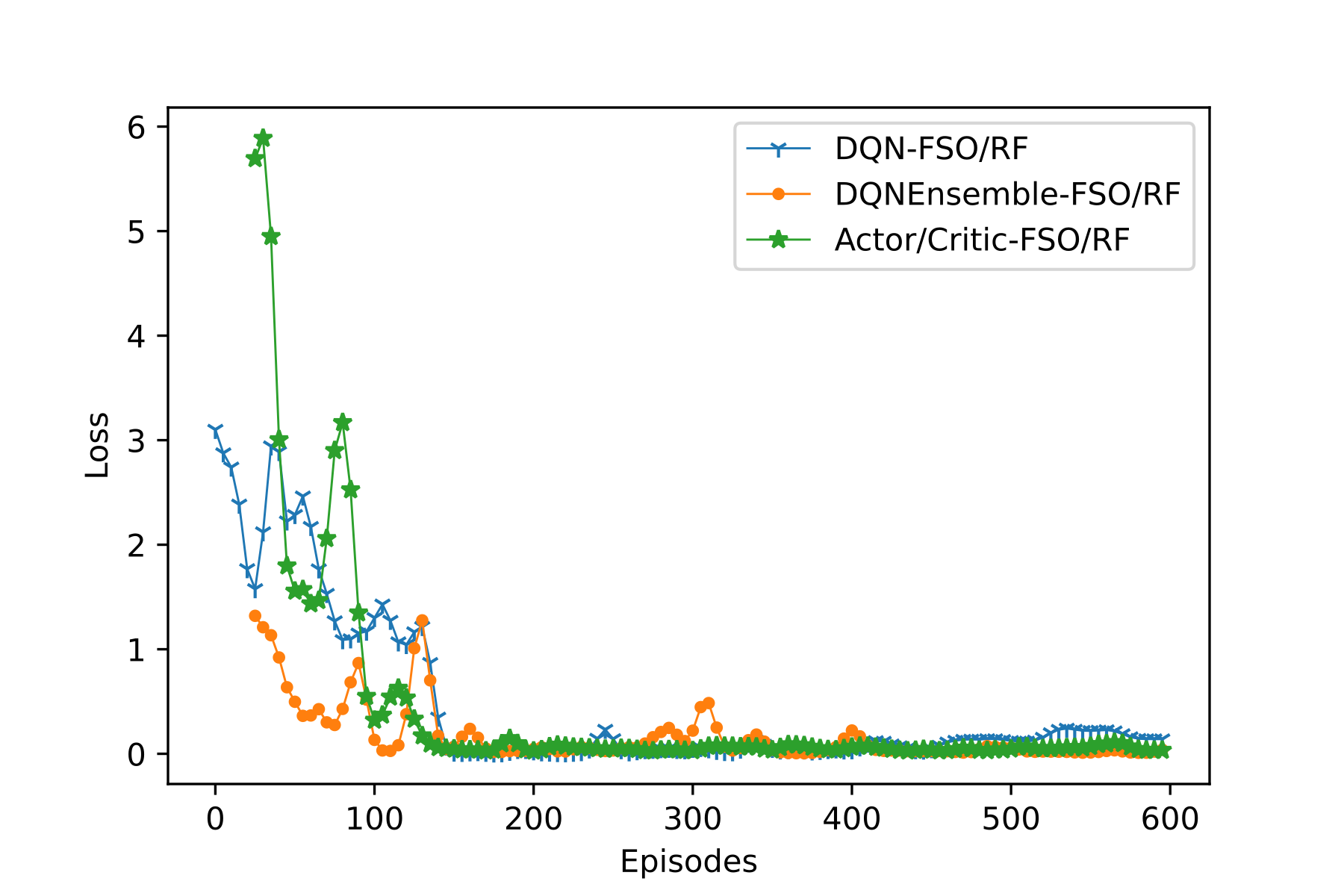}  
  
  \caption{Loss comparison of DQNEnsemble-FSO/RF,  DQN-FSO/RF, and Actor/Critic-FSO/RF.} \label{Lossepisode} 
\end{figure}

Figure \ref{Lossepisode} presents the loss results versus the number of episodes during the training of DQNEnsemble-FSO/RF,  DQN-FSO/RF, and Actor/Critic-FSO/RF. The figure shows the training loss for all the agents during the first training episodes. This is mainly attributed to the fact that the agents have not acquired sufficient information for the FSO/RF environment. However, owing to the deep neural networks, the loss drops rapidly, and for all the DRL agents, it stabilizes or converges around $100$ episodes, thereby indicating the capability of all these agents to adapt to the FSO/RF environment dynamics. The results show that for all three DRL agents, the loss degrades to almost zero with an increase in the number of episodes. It is evident from Figure \ref{Lossepisode} that DQNEnsemble-FSO/RF agent shows lesser loss value in contrast to  DQN-FSO/RF showing the success of DQNEnsemble-FSO/RF for FSO and RF switching. On the contrary, the Actor/Critic-FSO/RF  agent has the highest loss. This trend of the results demonstrates that all the DRL agents converge with an increase in the number of training episodes. The results also imply that all DRL agents converge within a reasonable number of training episodes. Actor/Critic-FSO/RF  converges faster as compared to DQNEnsemble-FSO/RF and DQN-FSO/RF.

\begin{figure}[ht] 
    \includegraphics[width=.55\textwidth]{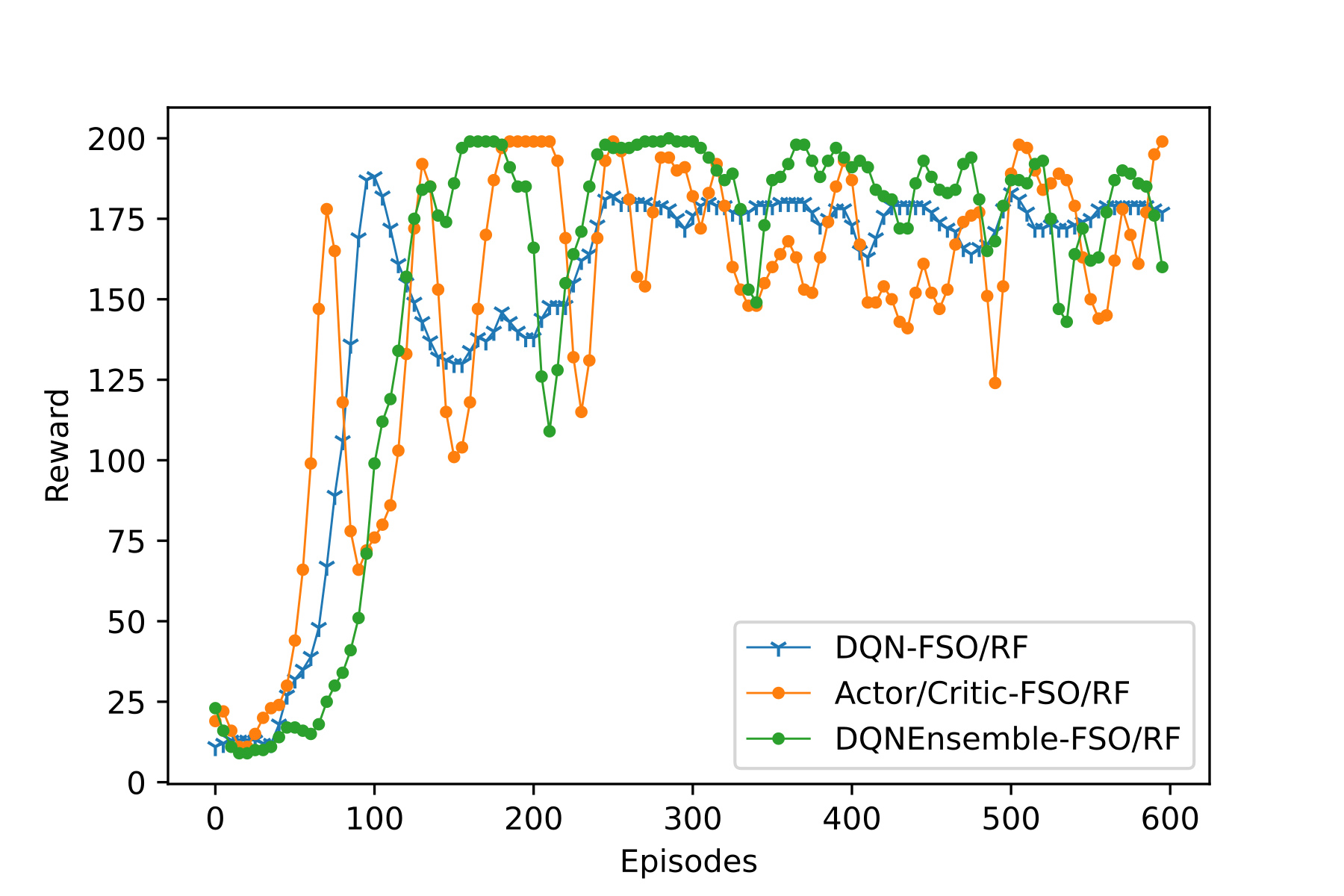}  
  
  \caption{Reward comparison of DQNEnsemble-FSO/RF,  DQN-FSO/RF, and Actor/Critic-FSO/RF .} \label{Rewardepisode} 
\end{figure}

Figure \ref{Rewardepisode} presents the average reward observed by all the DRL agents with a varying number of training episodes. For all the DRL agents,  increasing the number of episodes
leads to higher rewards and ultimately convergence. The figure shows that for the first $50$ training episodes, the reward for all the DRL agents is low. However, increasing the number of episodes leads to convergence with higher rewards. The reward results show that for DQNEnsemble-FSO/RF and DQN-FSO/RF, the average reward is highest after $100$ episodes.

Actor/Critic-FSO/RF agent achieves maximum average reward after $200$ episodes indicating the success of switching between RF and FSO links. This increase in average reward with the number of episodes indicates the convergence of DRL agents after a reasonable number of training episodes. Although, Actor/Critic-FSO/RF  achieves the highest possible reward between $100$ and $200$ episodes. However, instability in training results in some variance of the average reward over $300$ to $600$ episodes.
It can be observed that the DQNEnsemble-FSO/RF converges to a higher average reward than the DQN-FSO/RF and  Actor/Critic-FSO/RF. Both DQN-FSO/RF and  Actor/Critic-FSO/RF  achieve higher rewards with fewer training episodes compared to DQNEnsemble-FSO/RF, indicating faster learning. Further, in contrast to high variance of the average reward over multiple episodes for Actor/Critic-FSO/RF, DQN-FSO/RF, and DQNEnsemble-FSO/RF demonstrates stability in training and therefore less average reward variance over a longer number of episodes. Overall average reward results show that DQNEnsemble-FSO/RF is superior to that of DQN-FSO/RF and  Actor/Critic-FSO/RF, respectively, indicating the effectiveness of DQNEnsemble-FSO/RF learning due to a decrease in overestimation error of Q-value for FSO and RF switching.

\begin{figure}[ht] 
    \includegraphics[width=.55\textwidth]{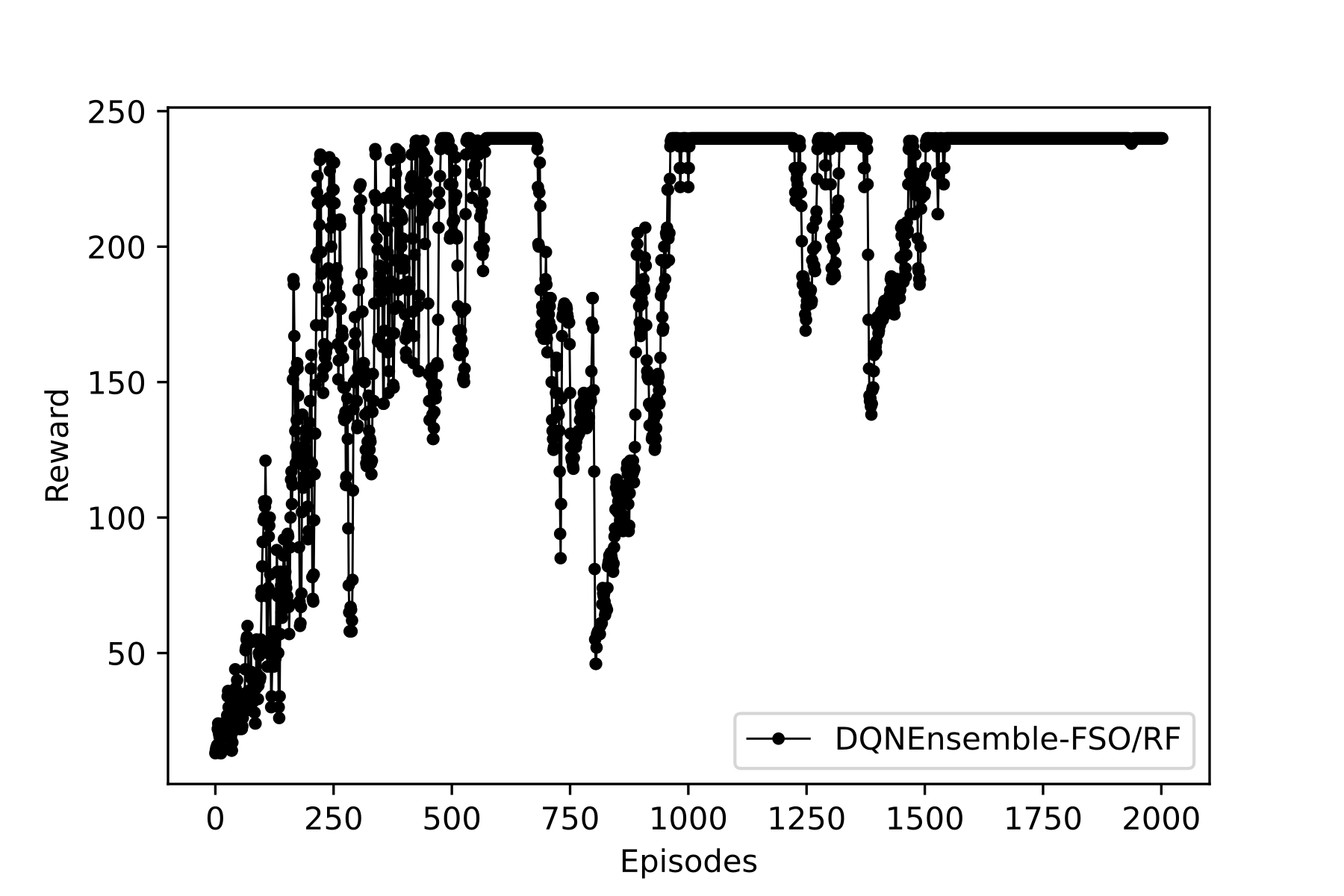}  
  
  \caption{Reward of DQNEnsemble-FSO/RF.} \label{DQNEnsemnleReward} 
\end{figure}

Figure \ref{DQNEnsemnleReward} shows the mean reward for the DQNEnsemble-FSO/RF after interacting with the environment for episodes varying from 250 to 2000. The DQNEnsemble-FSO/RF achieves convergence of approximately 1500 episodes. It can be observed that in the early episodes, the reward value is low due to the limited learning. As the number of training episodes increases, the DQNEnsemble-FSO/RF gradually evolves, and reward increases. The average reward per episode is significantly improved after 1000 episodes. After about 1500 training episodes, the reward flattens out smoothly, indicating DQNEnsemble-FSO/RF's ability for successful FSO and RF link switching.

	\begin{figure}[ht] 
    \includegraphics[width=.55\textwidth]{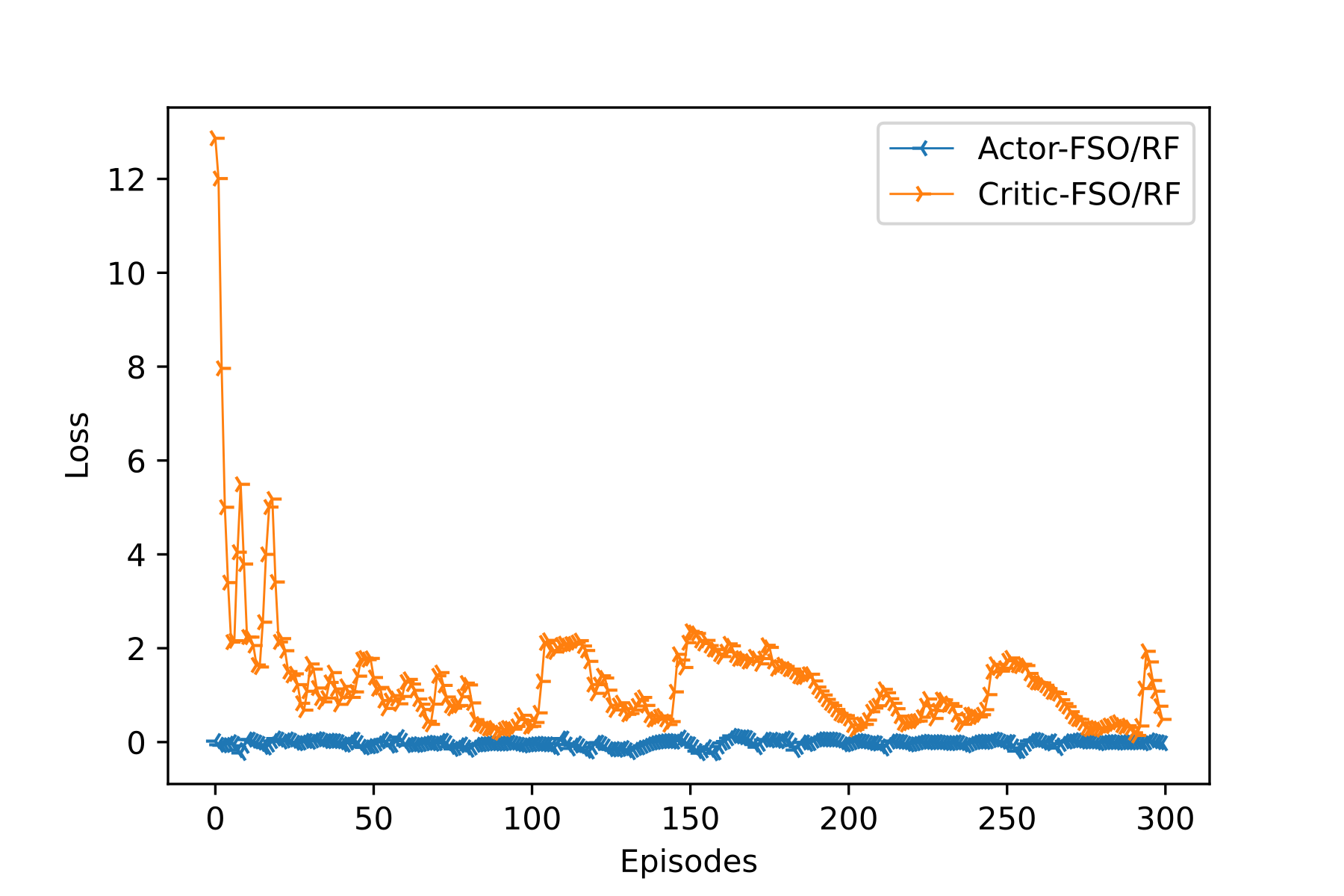}  
  
  \caption{Loss of actor versus critic in Actor/Critic-FSO/RF.} \label{Losscritic} 
\end{figure}

Figure \ref{Losscritic} shows the loss value variation of the critic and actor neural networks for the Actor/Critic-FSO/RF  agent. It can be seen from the figure that the average loss value of both the actor and critic decreases during the training process, which indicates that both the actor and critic reduce the error due to overestimations and helps the Q-learning. The loss curves show that the actor loss shows a dramatic decrease at the beginning of the training process, then the reduction gradually turns unstable. On the contrary, the critic loss is lower from the beginning and stays stable as the training continues. As a consequence, the action-value of the critic neural network achieves a higher reward. This, however, does not guarantee an optimal learned policy as the Actor/Critic-FSO/RF may overfit.

\begin{figure}[ht] 
    \includegraphics[width=.55\textwidth]{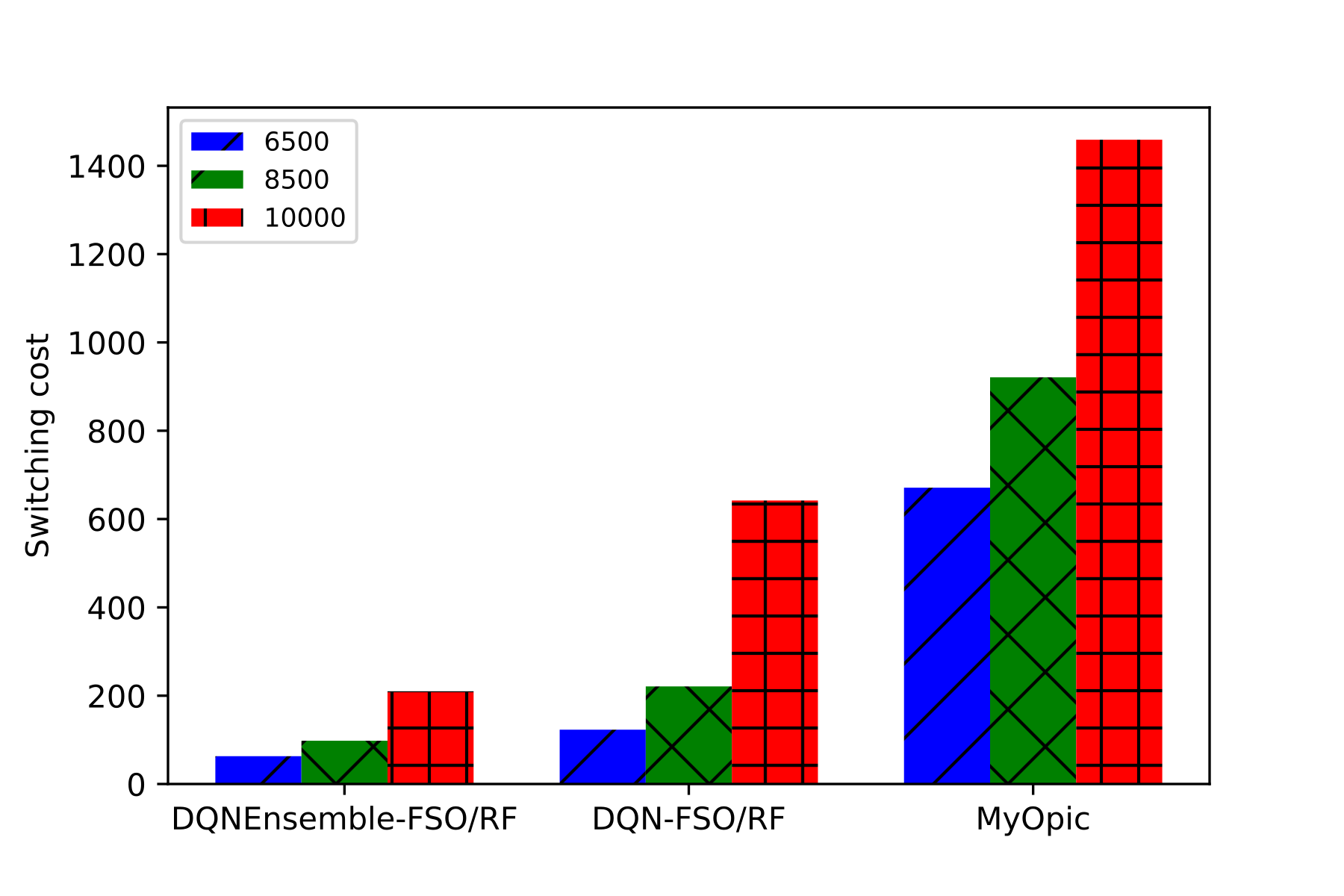}  
  
  \caption{Switching cost comparison.} \label{DQNswitchcost} 
\end{figure}

For DQNEnsemble-FSO/RF,DQN-FSO/RF, and MyOpic , we evaluate the switching cost for $6500$, $8500$, and $10,000$ episodes. During the training process, the target network policy is synchronized with the action-value policy using the consensus of features of $10$ threads.
As shown in Figure \ref{DQNswitchcost}, MyOpic with known $P$, i.e, 0.5 follows a round-robin fashion to switch policy and has the highest switching cost for all the episodes. MyOpic cannot use correlation among FSO, RF, and environment for policy switching.  DQN-FSO/RF shows a significantly lower switching cost than MyOpic. DQN-FSO/RF agent can learn the FSO/RF system dynamics including the correlation between FSO, RF, and environment, i.e., atmospheric conditions. The learned policy switching improves accumulated DQN-FSO/RF, thereby improving the FSO/RF system performance. In contrast to MyOpic and DQN-FSO/RF, DQNEnsemble-FSO/RF drastically reduces the policy switches that represent low switching cost suitable for FSO/RF hybrid systems operating under stable environments, such as atmospheric conditions. DQNEnsemble-FSO/RF's consensus criterion for policy switching achieves better performance with the minimal switching cost for $6500$, $8500$, and $10,000$ episodes. The consensus criterion switches the action-value policy decreases with an increase in the number of episodes. It results in a significant switching cost reduction compared to MyOpic and DQN-FSO/RF and remains more robust than MyOpic and  DQN-FSO/RF.

\begin{figure}[ht] 
    \includegraphics[width=.55\textwidth]{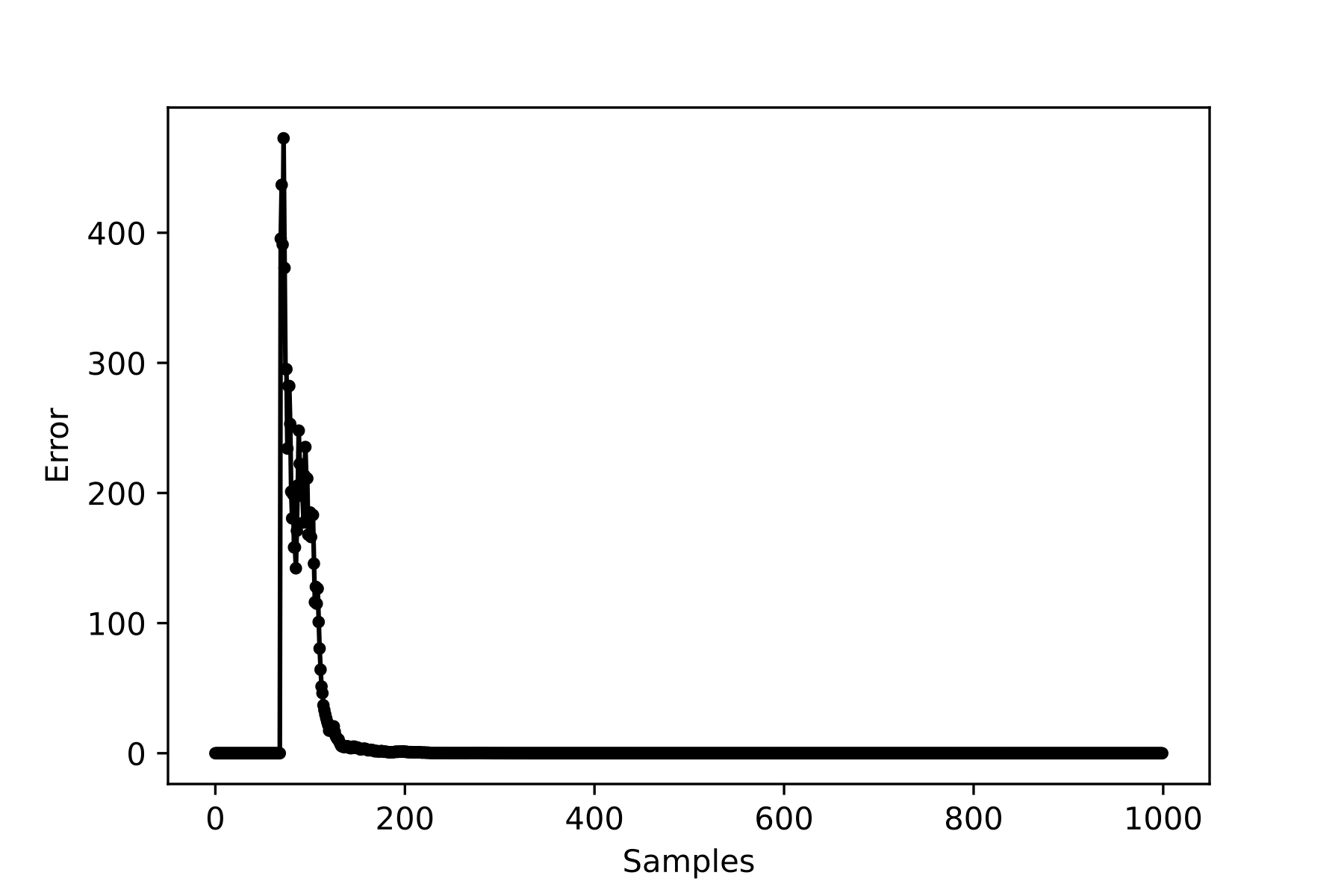}  
  
  \caption{DQNEnsemble-FSO/RF first transition.} \label{DQNEnsemblefirst} 
\end{figure}

\begin{figure}[ht] 
    \includegraphics[width=.55\textwidth]{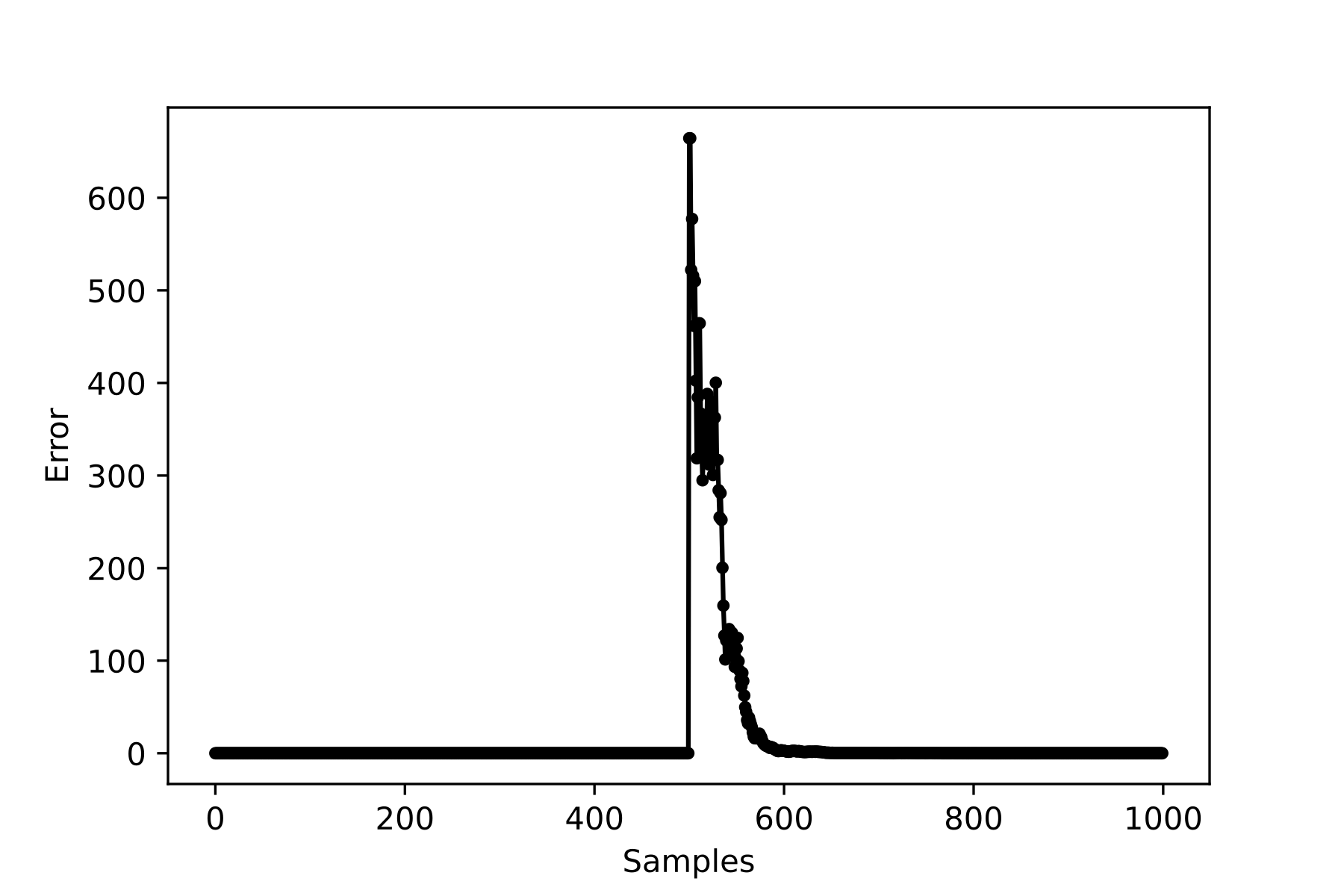}  
  
  \caption{DQNEnsemble-FSO/RF last transition.} \label{DQNEnsemblelast} 
\end{figure}

\begin{figure}[ht] 
    \includegraphics[width=.55\textwidth]{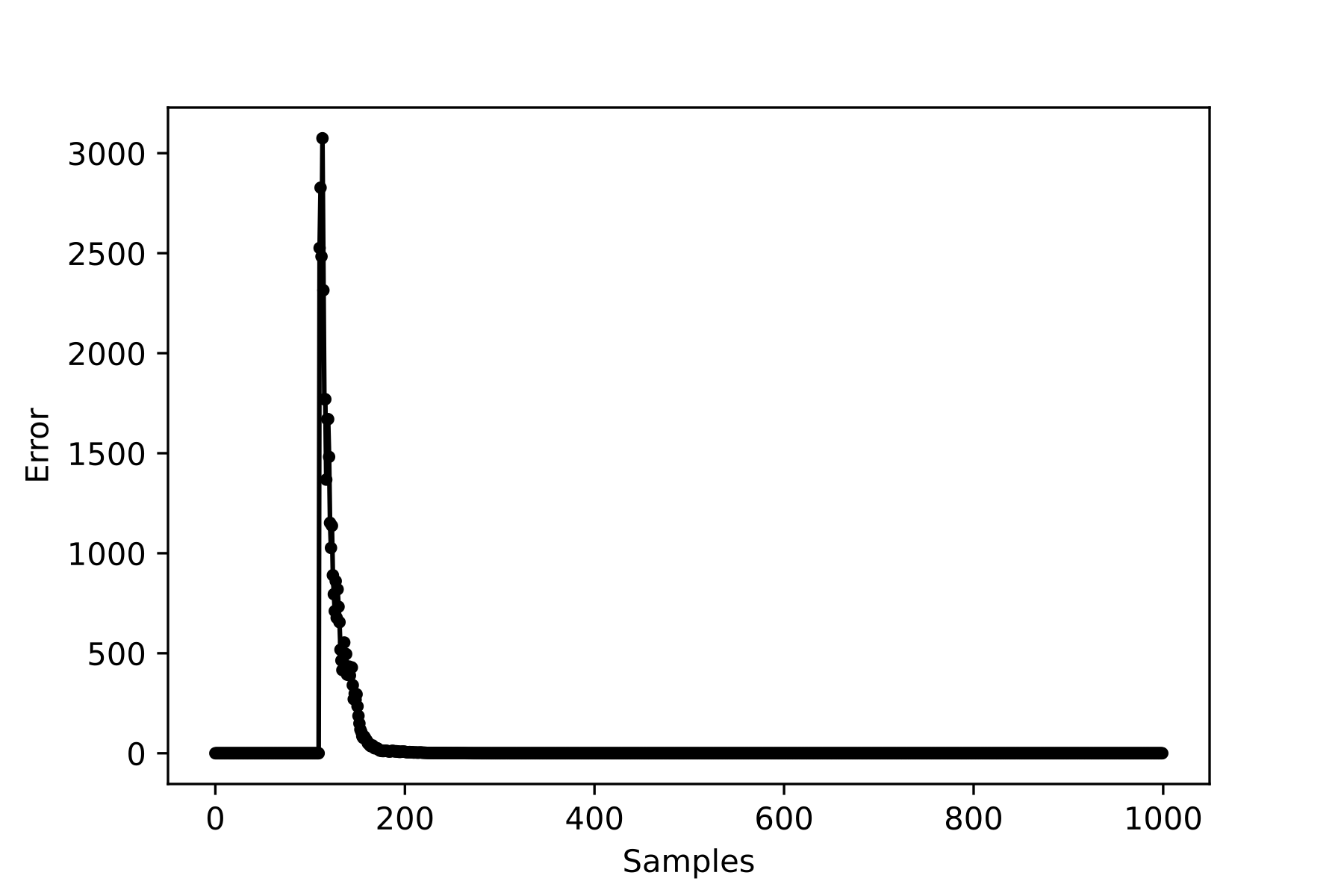}  
  
  \caption{DQN-FSO/RF first transition.} \label{DQNfirst} 
\end{figure}

\begin{figure}[ht] 
    \includegraphics[width=.55\textwidth]{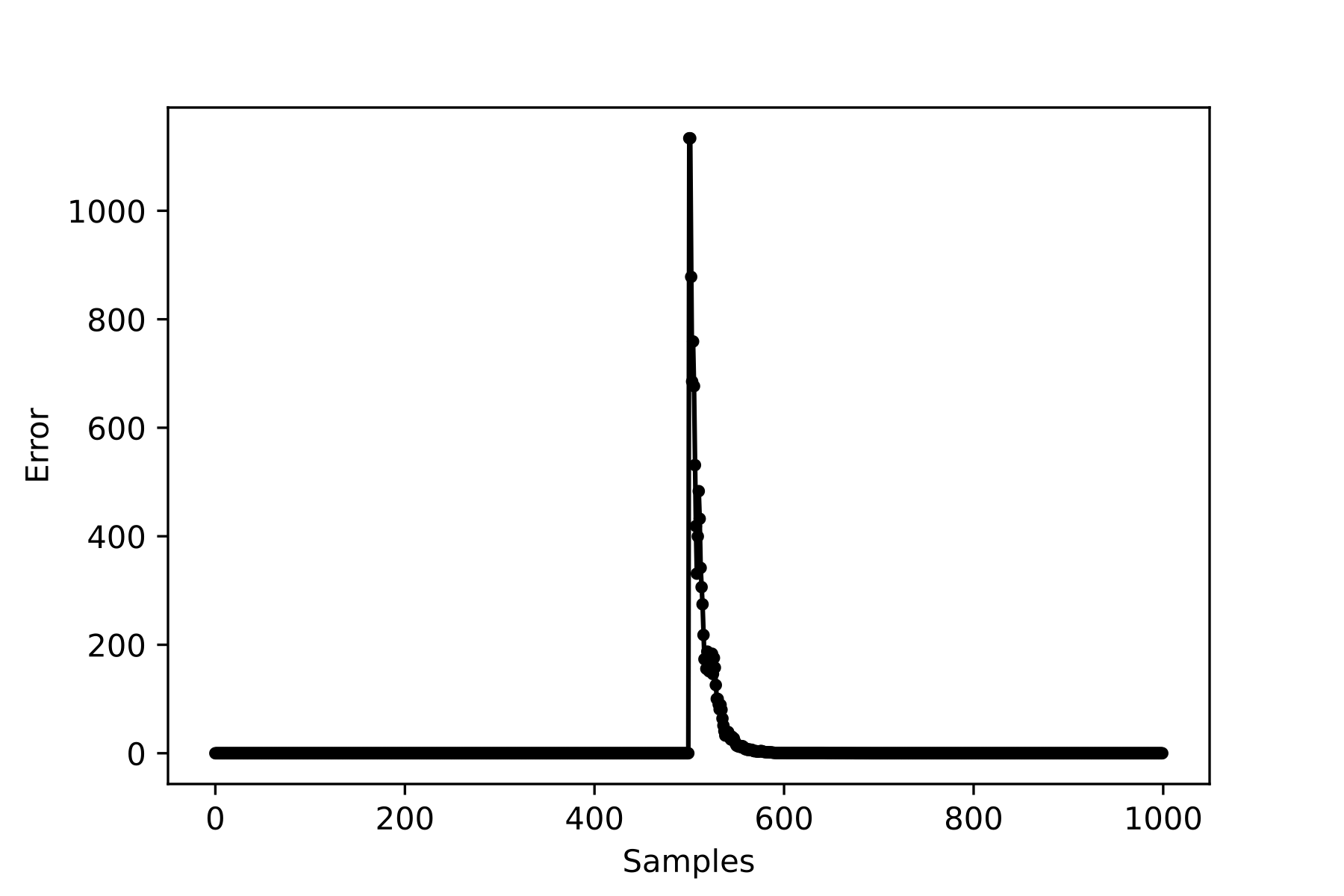}  
  
  \caption{DQN-FSO/RF last transition.} \label{DQNlast} 
\end{figure}

Figure \ref{DQNEnsemblefirst} investigates the first transition of the DQNEnsemble-FSO/RF execution with a varying number of samples.  It can be observed from the Figure that DQNEnsemble-FSO/RF's mean error converges faster between 0 and 40 steps. The plot considers the error calculation over the initial $100$ steps to calculate the mean error. Figure \ref{DQNEnsemblelast} plots the error of the DQNEnsemble-FSO/RF for the last transition for a sample captured during the agent execution. This error value for this transition is used to calculate the means to analyze DQNEnsemble-FSO/RF's learning performance. Similar to DQNEnsemble-FSO/RF, Figure \ref{DQNfirst} and Figure \ref{DQNlast} show the first and the last transition of the DQN-FSO/RF agent, respectively. These samples represent the smoothest transitions during the agent execution, demonstrating faster convergence.

\begin{figure}[ht] 
    \includegraphics[width=.55\textwidth]{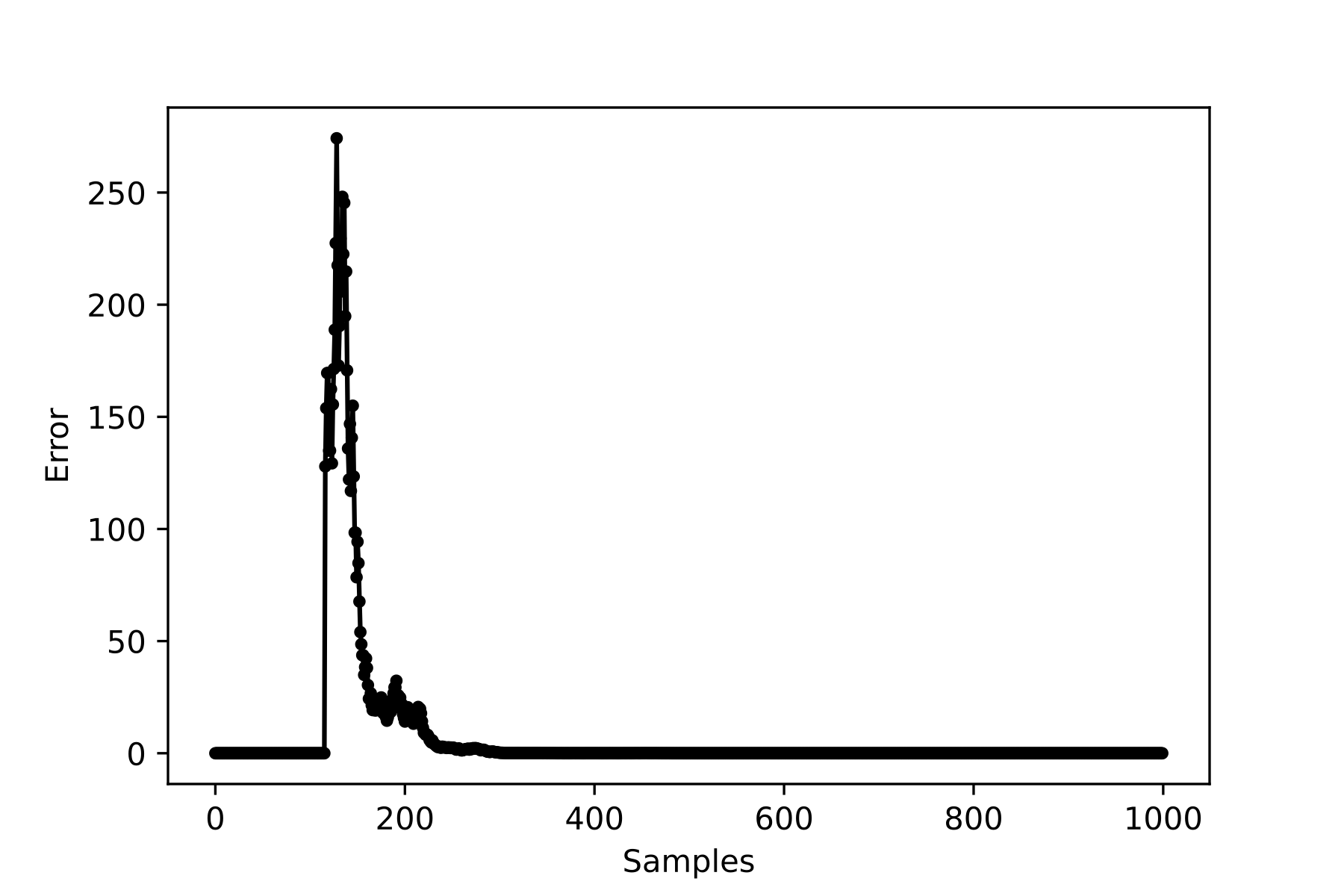}  
  
  \caption{Actor/Critic-FSO/RF first transition.} \label{ActorCriticfirst} 
\end{figure}

\begin{figure}[ht] 
    \includegraphics[width=.55\textwidth]{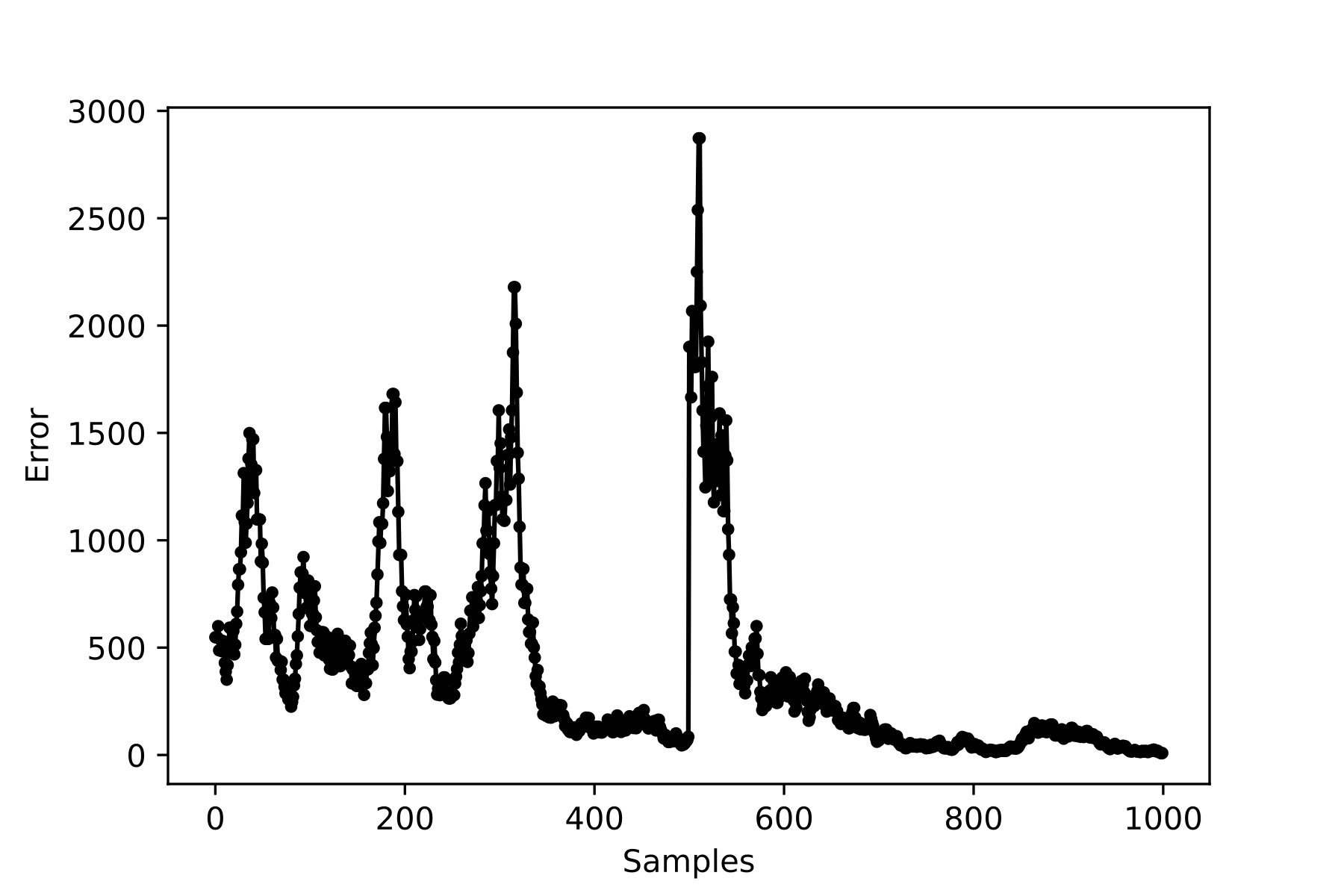}  
  
  \caption{Actor/Critic-FSO/RF last transition.} \label{ActorCriticlast} 
\end{figure}

Figure \ref{ActorCriticfirst} and Figure \ref{ActorCriticlast} plots the behaviour Actor/Critic-FSO/RF for the first and last transitions, respectively. It is evident from the figure that the Actor/Critic-FSO/RF does not converge in contrast to DQNEnsemble-FSO/RF and DQN-FSO/RF. Further, for the first transition, its mean error remains significantly higher compared to DQNEnsemble-FSO/RF and DQN-FSO/RF, resulting in slow learning. The last transition, however, demonstrates a significantly lower mean error and is comparable to DQNEnsemble-FSO/RF and DQN-FSO/RF.

\begin{figure}[ht] 
    \includegraphics[width=.55\textwidth]{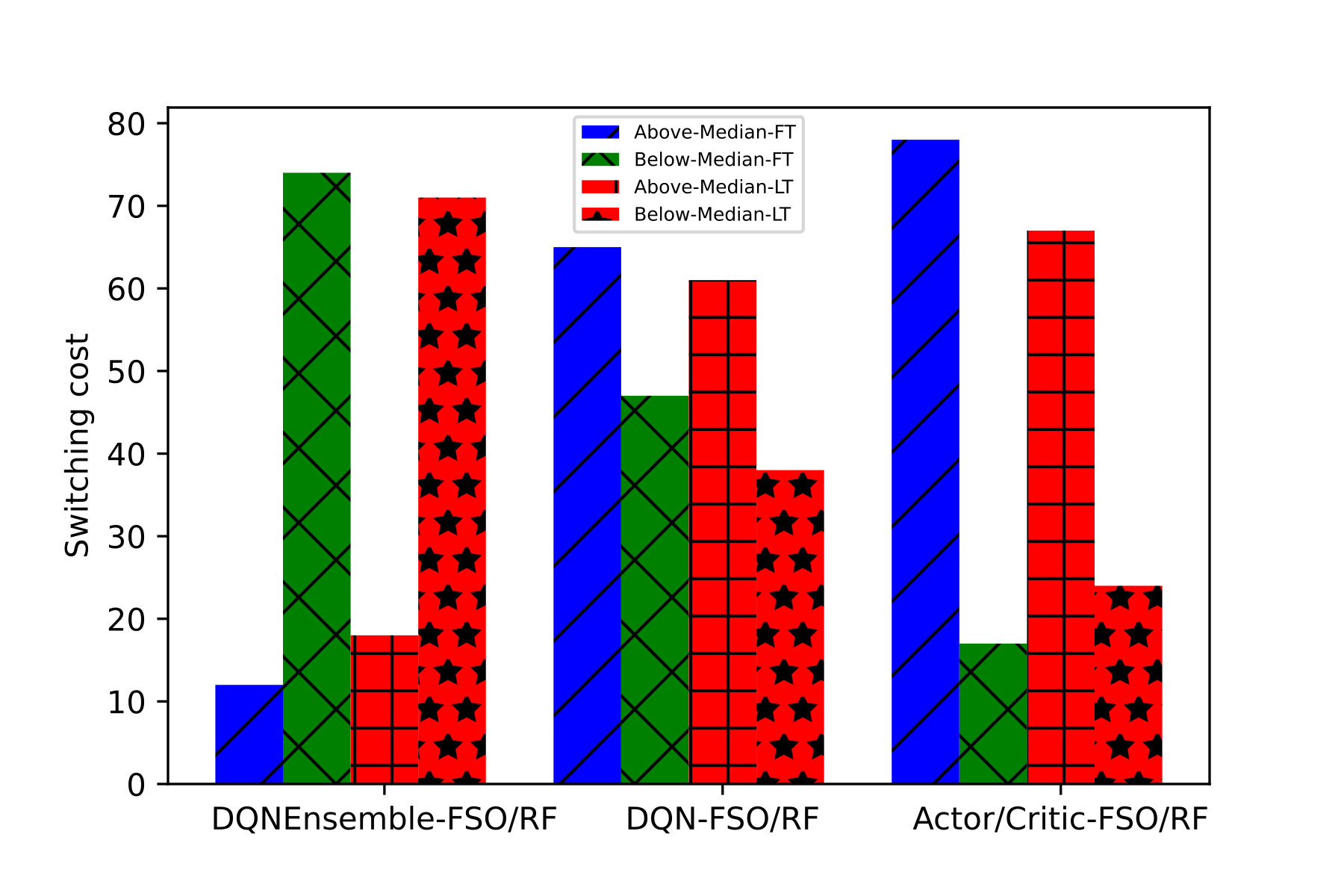}  
  
  \caption{DQNEnsemble-FSO/RF stability.} \label{DQNstability} 
\end{figure}

Figure \ref{DQNstability} demonstrates the stability of DQNEnsemble-FSO/RF, DQN-FSO/RF, and Actor/Critic-FSO/RF. The stability, here, represents, the frequency an agent deviates from the median error for both the first and last transitions. As discussed, earlier, the last transition of each agent is used to calculate the median value as it indicates an agent's best performance. It can be observed from the Figure that DQNEnsemble-FSO/RF significantly deviates less from the median for the last transition in contrast to  DQN-FSO/RF and Actor/Critic-FSO/RF. This proves DQNEnsemble-FSO/RF's stable learning for the first transition. Its count for the deviation from the median value for the last transition is comparable to the other agents. This shows that DQNEnsemble-FSO/RF exhibits overall better stability for both the first and the last transition attributed to its consensus representation learning to update the deployed policy. The  Actor/Critic-FSO/RF  agent demonstrates the least stability among all the agents, therefore, has slow learning performance.

\subsection{RSSI versus Visibility Analysis of DQNEnsemble-FSO/RF }

\begin{figure}[ht] 
    \includegraphics[width=.55\textwidth]{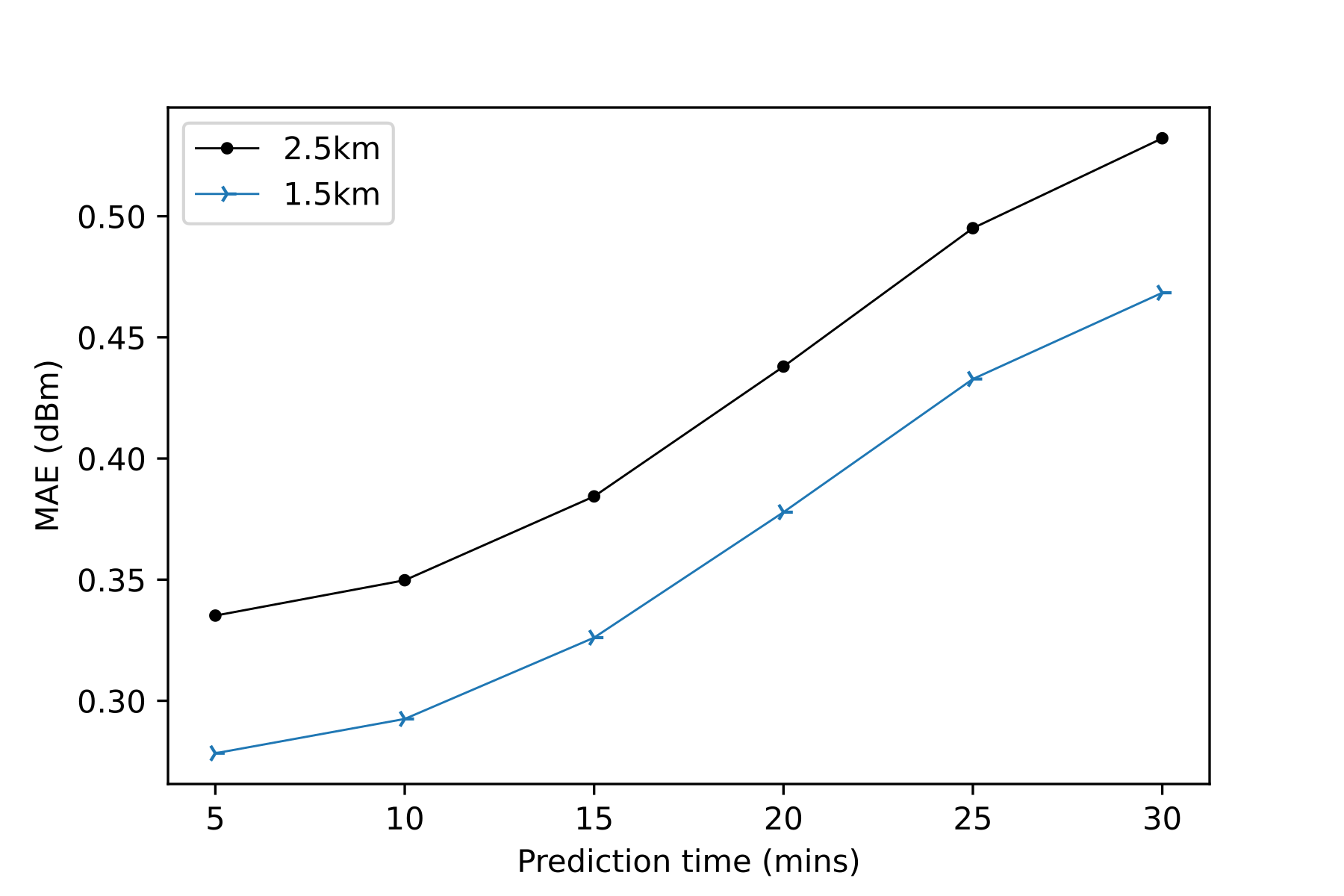}  
  
  \caption{MAE (dBm) vs. prediction time for DQNEnsemble-FSO/RF.} \label{MAE} 
\end{figure}

This section presents a performance analysis of the DQNEnsemble-FSO/RF agent under the FSO/RF environment with various visibility ranges of $5$km to $30$km. Figure \ref{MAE} shows MAE of the received signal strength indicator (RSSI) with various prediction steps. The results are taken over the time interval of 5 minutes, therefore each prediction step corresponds to 5 minutes. The figure shows that an increase in the number of prediction steps increases errors. This is mainly attributed to the higher weights associated with the RSSI values from the data collected over time affecting the accuracy of predicted results. A reduction in visibility of FSO/RF environment causes fluctuations in RSSI monitored by the DQNEnsemble-FSO/RF agent affecting the accuracy of prediction.

It can be seen from the Figure  \ref{MAE} that during 5 minutes RSSI values observations by the DQNEnsemble-FSO/RF, the MAE values at 1.5 km and 2.5 km are 0.254dBm  and 0.341dBm, respectively. Similarly, when the predictions are taken over 25 minutes, these values at 1.5 km and 2.5 km are increased to 0.423dBm and 0.48dBm, respectively. From these results, the MAE rate for both the 1.5 km and 2.5 km are 83.1\% and 86.1\%, respectively. It is evident from the results that the increase in prediction error rate is lower and acceptable level with a fewer number of prediction steps. The randomness in observed RSSI due to visibility fluctuations is not a true linear function that is significantly impacted by the agent DQNEnsemble-FSO/RF learning.

\begin{figure}[ht] 
    \includegraphics[width=.55\textwidth]{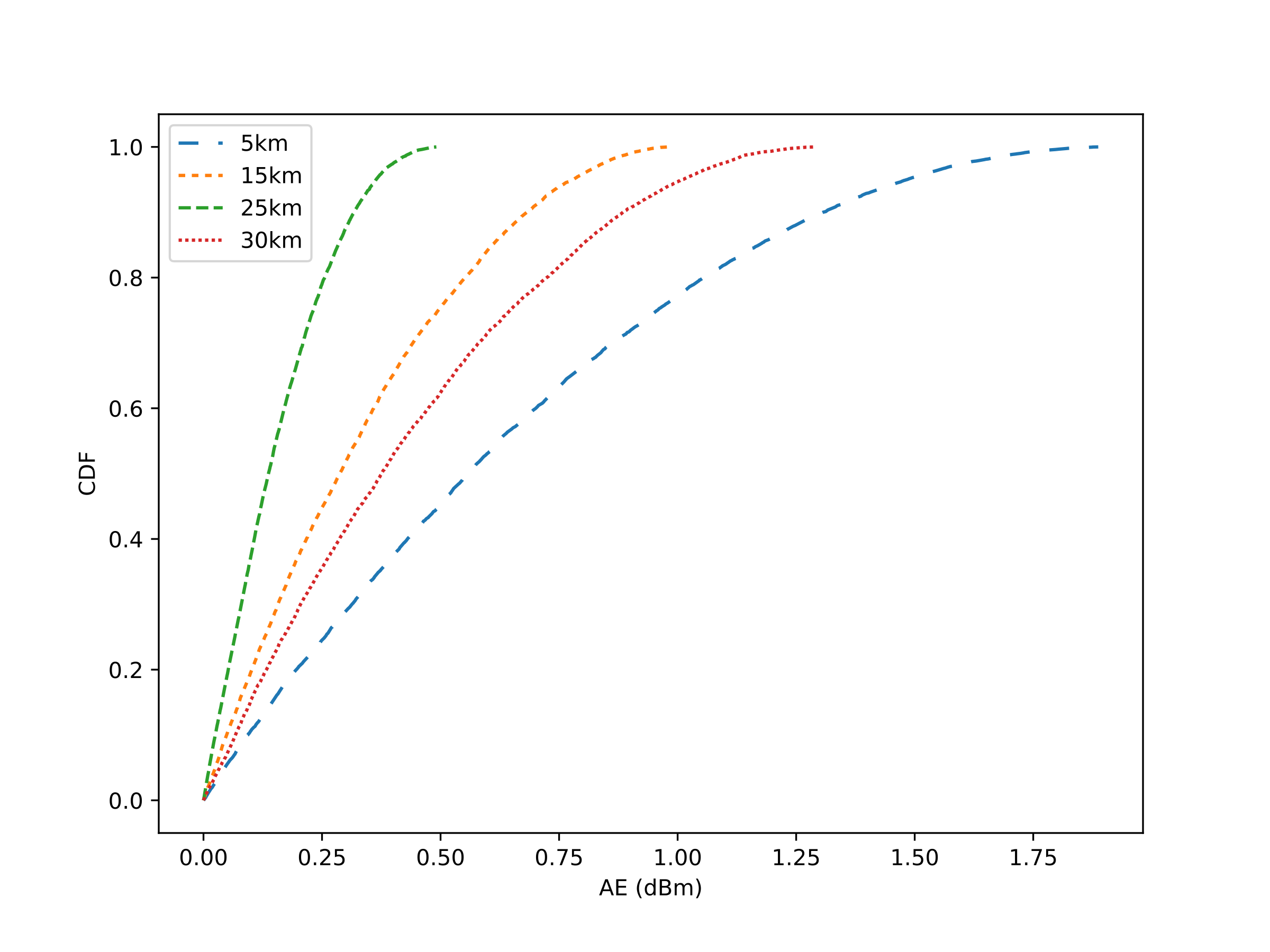}  
  
  \caption{CDF vs. AE for various visbility ranges.} \label{visibilitycomp} 
\end{figure}

To understand the probabilistic distribution of mean absolute error by DQNEnsemble-FSO/RF under visibility values of 5km to 30km, we plot the cumulative distribution function (CDF) of the absolute error (AE) in Figures \ref{visibilitycomp}. AE is computed using the predicted RSSI by the DQNEnsemble-FSO/RF  and comparing it with the true RSSI. Figure shows that AE values lower than 0.6 accounts for 50\% of errors for 5km. On the contrary,  Figure shows that AE lower than less than 0.6 contributes to more than 70\%. Figure \ref{visibilitycomp} shows that the percentage increases to above 90\% for the AE data values lower than 0.5. Finally, the AE data lower than 0.6 accounts for a maximum of 70\%. These results show that the prediction results of the  DQNEnsemble-FSO/RF increase with an increase in visibility ranges. This, however, maintains considerable prediction accuracy.

\section{Conclusion and Future Works} \label{conc}
To overcome the challenges of unknown weather dynamics, such as fog, dust, and sand storms, this work has applied the concept of ensemble consensus representation deep reinforcement learning for FSO/RF link switching for hybrid FSO/RF systems. The link switching optimization problem has been formulated to achieve the maximum long-term  FSO and RF  links utility as a whole while maximizing the link availability between the transmitter and receiver. Considering the non-convex and combinatorial characteristics of this optimization problem, we have applied actor-critic and DQN under a hybrid FSO/RF system with dynamic weather conditions. Compared with the actor-critic, the DQN algorithm achieves faster convergence. Further, to reduce the frequent switching of the deployed policy of DQN, we have proposed a consensus representation of DQN called DQNEnsemble-FSO/RF for FSO and RF link switching. Experiment results show that the DQNEnsemble-FSO/RF outperformed DQN-FSO/RF, Actor/Critic-FSO/RF, and MyOpic with faster convergence while maintaining switching costs significantly low under time-varying weather scenarios. The results also prove the efficiency of DQNEnsemble-FSO/RF for predicting RSSI for FSO/RF link switching with minimum absolute error for various visibility ranges. We believe this work is the first step toward the application of DRL for the link switching problem with consideration of low switching cost for hybrid FSO/RF systems. One interesting direction is to design a deep Q-network algorithm with provable guarantees and generalization that can consider the switching cost for a larger state space compared to the states considered in this work.



%

\ifCLASSOPTIONcaptionsoff
  \newpage
\fi



%

\begin{IEEEbiography}[{\includegraphics[width=1in,height=1.25in,clip,keepaspectratio]{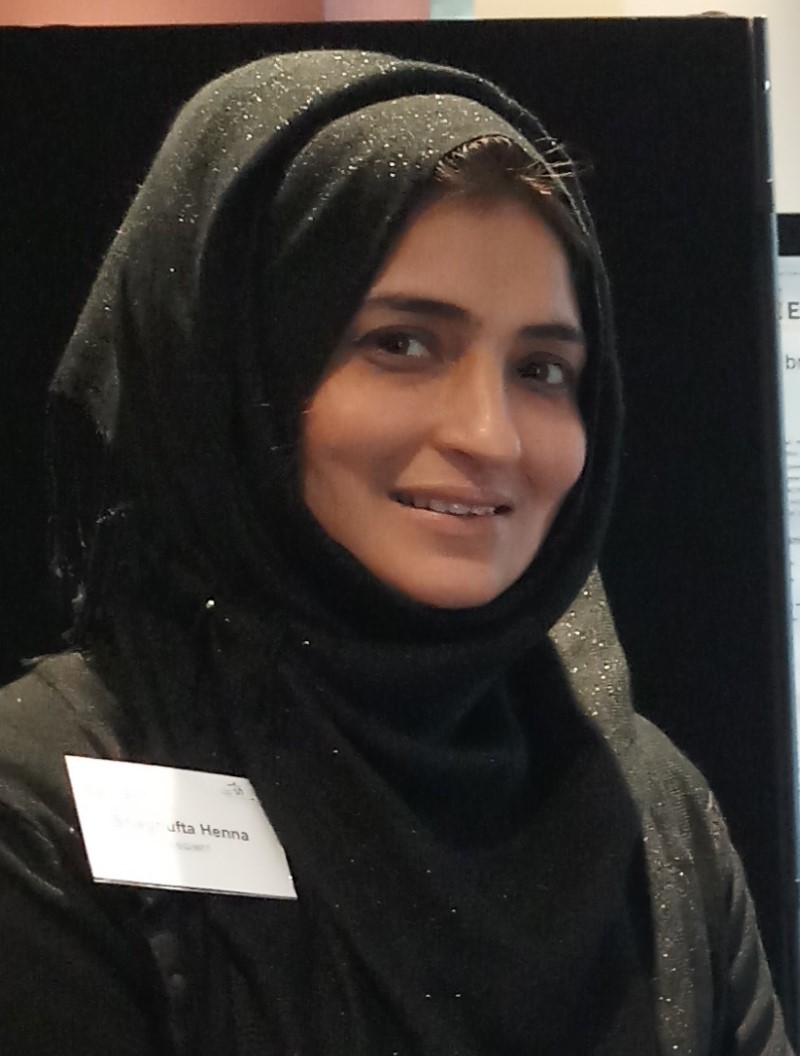}}]{Dr. Shagufta Henna} is lecturer with the Atlantic Technological University, Donegal, Ireland. She was a post-doctoral researcher with the  Waterford institute of technology, Waterford, Ireland from 2018 to 2019. She received her doctoral degree in Computer Science from the University of Leicester, UK in 2013. She is an Associate Editor for IEEE Access,EURASIP Journal on Wireless Communications and Networking, IEEE Future Directions, and Human-centric Computing and Information Sciences, Springer. Her current research
interests include edge computing, self-supervised learning/attention, representation learning, deep reinforcement learning, XAI, and AI-driven network optimizations.
\end{IEEEbiography}
\begin{IEEEbiography}[{\includegraphics[width=1in,height=1.25in,clip,keepaspectratio]{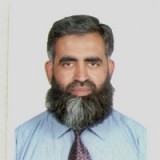}}]{Dr. Abid Ali Minhas }
is professor in the department of Computer Engineering, Al Yamamah University, Riyah, Saudi Arabia. He did his PhD in Wireless Sensor Networks from Graz University of Technology, Austria. He is the  author of a book, book chapters and published 75 research publications in International Journals/Conferences. He has experience of working in Funded 478 Projects. His research interest includes Wireless and wired communications, Wireless Sensor
Networks, FSO, Cybersecurity and hardware design.
\end{IEEEbiography}
\begin{IEEEbiographynophoto}{Dr. Muhammad Saeed Khan}
 received his M.Sc (Electronics) degree in 2004 from Quaid-E-Azam University, Islamabad, Pakistan. He did his PhD in Electrical Engineering from the
Graz University of Technology, Austria in 2012 in optical wireless communication.
 He is working as associate professor in the department of electrical
 engineering, Lahore Leads University, Lahore, Pakistan. His research
 interests are channel modeling, system design aspects and channel characterization  of Free-Space Optical Communications links. He has been involved in different EU projects like, SatNEx II, COST 291.
\end{IEEEbiographynophoto}

\begin{IEEEbiography}[{\includegraphics[width=1in,height=1.25in,clip,keepaspectratio]{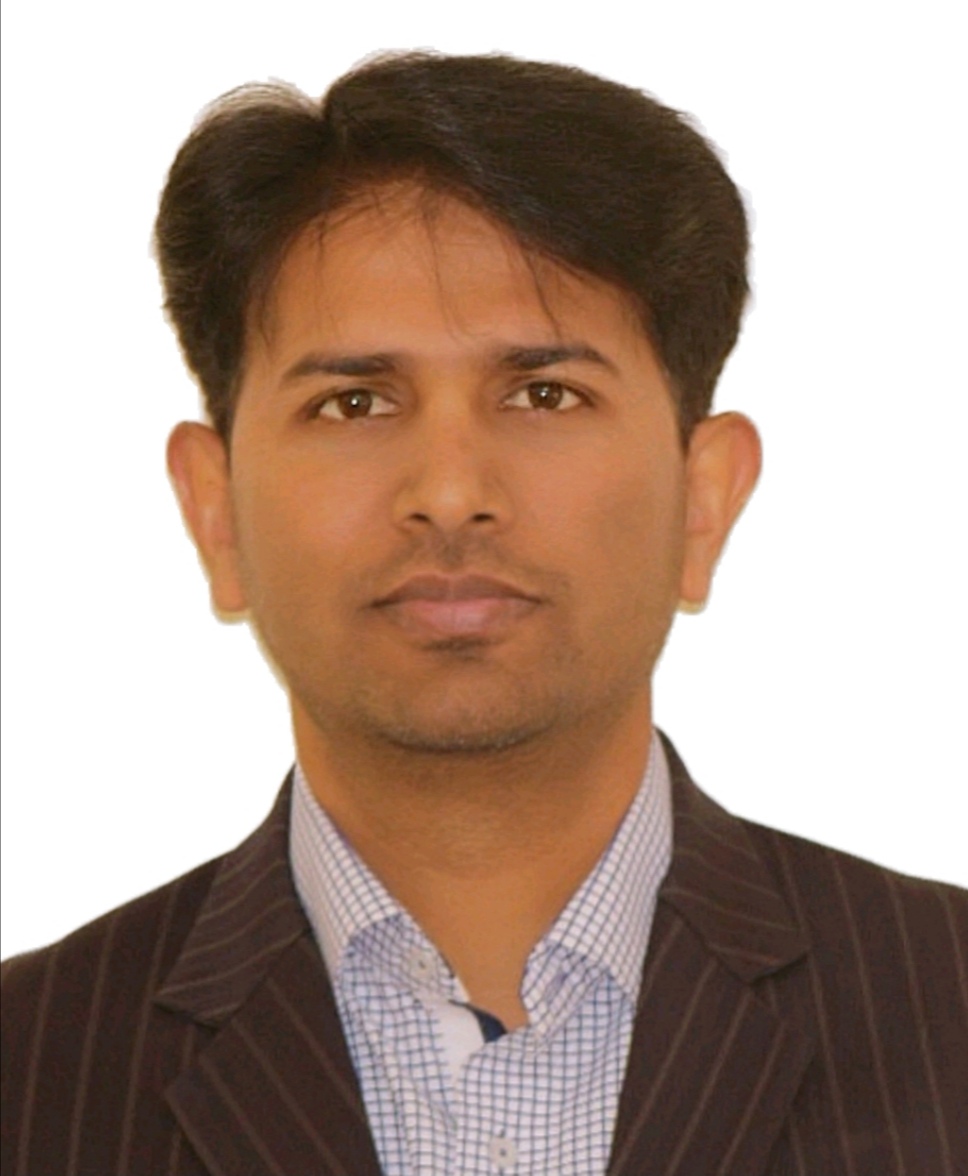}}]
{Dr. Muhammad Shahid Iqbal}
received MS degree in electrical engineering from
National University of Sciences and Technology (NUST), Islamabad, Pakistan in 2013 and
Ph.D. degree in electrical and electronics engineering from Koc University, Istanbul in 2021.
He has been a lecturer with the Department of Electrical Engineering, The University of
Lahore (2013-2014) and with the University of Gujrat(2014-2016). Currently, he is working as an Assistant Professor with the  Department of Electrical engineering National university of Technology,Islamabad, Pakistan. His research interests include wireless communication design, wireless powered communication networks, and sensor networks
\end{IEEEbiography}




\end{document}